%% file: main.tex
\documentclass[sigconf]{acmart}

\usepackage{epsfig}
\usepackage{graphicx}
\usepackage{amsmath,amsfonts} 
\usepackage{hyperref} 
\usepackage{tabulary,color}
\usepackage{tabularx}
\usepackage{multirow}
\usepackage{enumitem} 
\usepackage{array} 
\usepackage{url} 
\usepackage[caption=false,font=footnotesize]{subfig}
\usepackage{pifont}

\newlength\savewidth\newcommand\shline{\noalign{\global\savewidth\arrayrulewidth\global\arrayrulewidth 1pt}\hline\noalign{\global\arrayrulewidth\savewidth}}

\renewcommand\footnotetextcopyrightpermission[1]{} 

\AtBeginDocument{%
  \providecommand\BibTeX{{%
    \normalfont B\kern-0.5em{\scshape i\kern-0.25em b}\kern-0.8em\TeX}}}

\copyrightyear{2021}
\acmYear{2021}
\setcopyright{acmcopyright}
\acmConference[MM '21] {Proceedings of the 29th ACM Int'l Conference on Multimedia}{October 20--24, 2021}{Virtual Event, China.}
\acmBooktitle{Proceedings of the 29th ACM Int'l Conference on Multimedia (MM '21), Oct. 20--24, 2021, Virtual Event, China}
\acmPrice{15.00}
\acmISBN{978-1-4503-8651-7/21/10}
\acmDOI{10.1145/000000.000000}

\begin{document}
\fancyhead{}

\title{Joint Implicit Image Function for Guided Depth Super-Resolution}

\author{Jiaxiang Tang, Xiaokang Chen, Gang Zeng}
\affiliation{
  \institution{Key Laboratory of Machine Perception (MOE), Peking University}
  \city{Beijing}
  \country{China}
}
\email{{tjx, pkucxk, zeng}@pku.edu.cn}

\begin{abstract}
Guided depth super-resolution is a practical task where a low-resolution and noisy input depth map is restored to a high-resolution version, with the help of a high-resolution RGB guide image.
Existing methods usually view this task as a generalized guided filtering problem that relies on designing explicit filters and objective functions, or a dense regression problem that directly predicts the target image via deep neural networks. 
These methods suffer from either model capability or interpretability.
Inspired by the recent progress in implicit neural representation, 
we propose to formulate the guided super-resolution as a neural implicit image interpolation problem, where we take the form of a general image interpolation but use a novel Joint Implicit Image Function (JIIF) representation to learn both the interpolation weights and values.
JIIF represents the target image domain with spatially distributed local latent codes extracted from the input image and the guide image, and uses a graph attention mechanism to learn the interpolation weights at the same time in one unified deep implicit function.
We demonstrate the effectiveness of our JIIF representation on guided depth super-resolution task, significantly outperforming state-of-the-art methods on three public benchmarks.
Code can be found at \url{https://git.io/JC2sU}.
\end{abstract}



\maketitle

\section{Introduction}
\input{introduction}

\section{Related Work}
\input{related}

\section{Method}
\input{method}

\section{Experiments}
\input{experiments}

\section{Conclusion}
\input{conclusion}

\begin{acks}
This work is supported by the National Key Research and Development Program of China (2017YFB1002601), National Natural Science Foundation of China (61632003, 61375022, 61403005), Beijing Advanced Innovation Center for Intelligent Robots and Systems (2018IRS11), and PEK-SenseTime Joint Laboratory of Machine Vision.
\end{acks}

\bibliographystyle{ACM-Reference-Format}
\bibliography{references}

\end{document}

%% file: introduction.tex

\begin{figure}
    \centering
    \includegraphics[width=\linewidth]{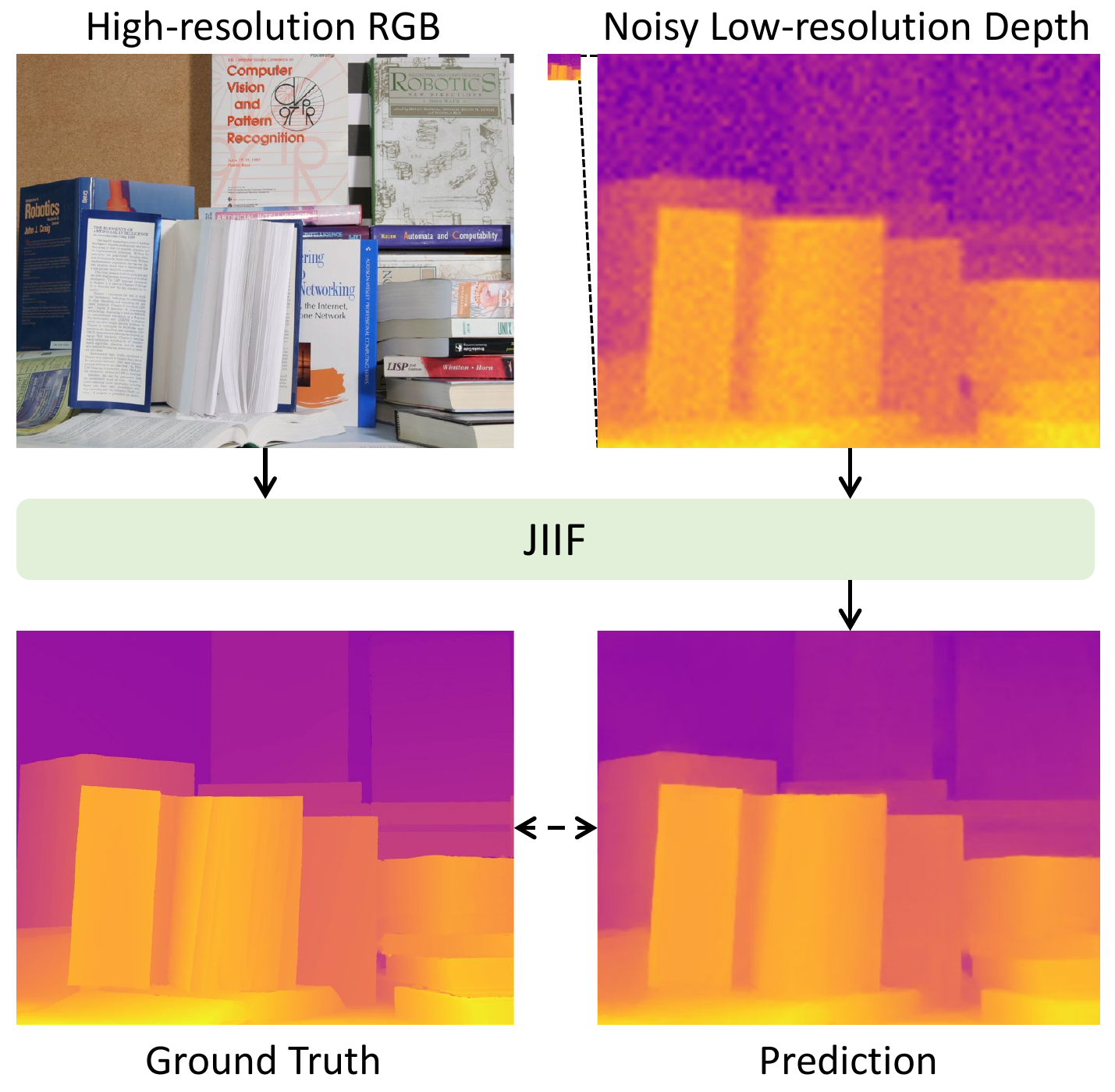}
    \caption{RGB guided noisy depth map super-resolution. 
    Our method predicts a high-resolution target depth map from a noisy and low-resolution input depth map with the guidance from a high-resolution RGB image.
    The low-resolution depth map is up-sampled with bicubic interpolation for better visualization.
    }
    \label{fig:teaser}
    \vspace{-0.5cm}
\end{figure}

Depth maps have been widely used as a basic element in various computer vision tasks, such as semantic segmentation~\cite{weder2020routedfusion,chen2020bi,gupta2014learning,xing20192,xing2019coupling} and 3D reconstruction~\cite{song2017semantic,chen20203d,chen2020real}.
With the geometric information in the depth maps, these tasks can be facilitated and better understood.
Despite the improvement of the depth sensors in recent years, high quality depth maps are still challenging to acquire. 
The acquired depth maps are usually low-quality due to the limitation of the sensors.
In the meantime, RGB cameras have evolved rapidly and can acquire high-quality RGB images with a comparatively low cost.
Hence, RGB guided depth super-resolution, where a high-resolution (HR) RGB image is used to guide the up-sampling process of a low-resolution (LR) depth input image, has become an important research topic.
As illustrated in Figure~\ref{fig:teaser}, the detailed structures in the RGB image can be used to avoid blurry edges and suppress noises when up-sampling depth maps, but this process is non-trivial due to the complicated nature of RGB images. 

Various methods have been proposed to extract the rich information from the RGB images to guide the depth super-resolution. 
The widely used bilateral filtering~\cite{Tomasi1998BilateralFF} and guided filtering~\cite{he2013guided} can be extended to solve this task, when we first up-sample the LR input image with a traditional interpolation method, e.g., bilinear or bicubic interpolation, to the same size as the HR guide image. 
These methods construct explicit filters that reflect structural information from the guide image.
However, a common drawback to these filtering based methods is that the filter is only dependent on the guide image, and it may transfer incorrect contents if the guide image is inconsistent with the input image (e.g., the RGB guide image is too dark to distinguish the structure).
Another popular way to model guided super-resolution is to treat it as a dense regression problem, and train a neural network to directly predict the HR target image from the LR input image and the HR guide image~\cite{li2016deep,Li2019JointIF,Hui2016DepthMS} in a supervised way. 
They first down-sample the HR targets to generate LR inputs, then learn to undo the down-sampling process.
With the strong capability of CNNs to extract features, these methods often outperform traditional methods.
However, the simple feature aggregation doesn't model the guide process explicitly, which lacks interpretability and may not generalize well to other datasets.

Inspired by the recent progress of implicit neural representation in 3D object/scene representation~\cite{Park2019DeepSDFLC,Jiang2020LocalIG,Rist2020SCSSnetLS} and image super-resolution~\cite{Chen2020LearningCI}, 
we revisit the guided super-resolution problem and view it from the perspective of implicit neural representation.
The idea of implicit neural representation is to use a deep implicit function (DIF) to map continuous coordinates to signals in a certain domain. 
To share knowledge across different input observations, an encoder is often used to extract latent codes from the input to make the DIF conditional to the current observation.
Thus, a scene/image can be represented by a set of local latent codes distributed in the coordinates of the input domain, which can be used in downstream tasks such as semantic segmentation~\cite{Rist2020SCSSnetLS} and super-resolution~\cite{Chen2020LearningCI}.
To make the output of DIF continuous, a weighted average of the predictions from several neighboring coordinates is usually calculated, which can be viewed as a neural implicit interpolation process.
However, these weights are usually empirical (e.g., distance-based) in previous work since there is no other prior knowledge between the query coordinate and the neighboring coordinates.
With the extra HR guide image in the guided super-resolution task, we can learn to extract this knowledge and learn the weights in a data-driven way.
We hypothesize that the guide image can benefit the learning of both interpolation weights and values, 
and propose to learn the interpolation weights via a graph attention mechanism.
Furthermore, we integrate the learning of weights and values into one unified DIF, which we call the joint implicit image function (JIIF) representation.

To summarize, the contributions of this paper are as follows:
\begin{itemize}
    \item We propose a novel joint implicit image function representation for guided image super-resolution, where the target image is represented by local latent codes from both the input image and the guide image.
    \item We learn interpolation weights at the same time via a graph attention mechanism, and integrate the learning of both interpolation weights and values into one unified representation.
    \item Our method outperforms existing methods by large margins on guided depth super-resolution tasks, and achieves state-of-the-art results on guided noisy depth super-resolution tasks.
\end{itemize}

%% file: related.tex

\subsection{Guided Super-Resolution}

\subsubsection{Filtering based methods}
Guided filtering aims to enhance a target image by applying a filter that is dependent on a guide image. 
Bilateral filter~\cite{Tomasi1998BilateralFF} is the starting work where the target image also serves as the guide image.
Later work includes joint bilateral filter~\cite{kopf2007joint}, guided filter~\cite{he2013guided} and weighted median filter~\cite{Ma2013ConstantTW}.
Guided filtering can be used for a variety of tasks such as image denoising, colourisation and stereo matching.
When extended to different sized target image and guide image, it can also be used for the guided super-resolution task.
We further distinguish these methods into two categories:
\textit{Local methods} first upsample the low-resolution target image with a traditional interpolation method,
then apply a local filter which is controlled by the guide image~\cite{kopf2007joint,Yang2007SpatialDepthSR} or both the target image and the guide image~\cite{Chan2008ANF}.
On the other hand, \textit{Global methods} formulate the filtering as an implicit energy minimization problem, and optimize values of all the pixels in the target image. 
This category includes Markov Random Field~\cite{diebel2006application} and its non-local means variant~\cite{park2011}. 
Variational inference with anisotropic total generalized variation prior~\cite{ferstl2013} and auto-regressive models~\cite{Yang2014ColorGuidedDR} are other types of global methods~\cite{Li2012JointED,Kiechle2013AJI}.
Some recent methods also combine the idea of guided filtering into the global optimization framework, such as the fast bilateral solver~\cite{barron2016fast} and the SD filter~\cite{ham2018robust}.

\subsubsection{Learning based methods}
Different from the previous unsupervised filtering based methods, learning based methods provide a data-driven and supervised way to solve the guided super-resolution problem by training neural networks.
Self-supervised super-resolution where the HR target image is first down-sampled to serve as the LR input has been explored by lots of methods~\cite{chang2004super,Dong2014LearningAD,Lai2017DeepLP,Zhang2020DeepUN,Zhang2018ResidualDN,Ledig2017PhotoRealisticSI,Chen2020LearningCI,Ye2020DepthSV}.
Guided super-resolution further introduces a HR guide image to direct the up-sampling process of the LR input.
Early work like Depth Multi-Scale Guided Network (DMSG)~\cite{Hui2016DepthMS}, Dynamic Guidance (DG)~\cite{Gu2017learning} and Deep Joint Filtering (DJF)~\cite{li2016deep,Li2019JointIF} starts to use CNNs to extract features and directly regress the target image.
Pixel-Adaptive Convolution (PAC)~\cite{su2019pixel} learns a spatially variant kernel to fuse the guide image features into the LR input.
Deformable Kernel Network (DKN)~\cite{Kim2021DeformableKN} draws ideas from both the explicit filtering based methods and learning based methods.
It uses a CNN to learn a set of sparsely chosen neighbors and the interpolation weights adaptively, then apply an explicit image filter to calculate the final prediction.
These methods either lack model interpretability for directly regressing the target, or rely on a simple image filter that cannot take full advantage of the guide image.
Instead, our method starts from the general form of image interpolation and equips it with the effective implicit neural representation, leading to both better performance and interpretability.

\subsection{Implicit Neural Representation}
Implicit neural representation uses a deep implicit function (DIF) to map coordinates to signals in a specific domain. 
A DIF is a continuous and differentiable function, usually parameterized by an MLP.
To make the DIF conditional to different input observations, a latent code extracted from the input is usually appended to the coordinate.
Recent research has demonstrated the potential of implicit neural representations for 3D single objects~\cite{Park2019DeepSDFLC,Xu2019DISNDI,Mescheder2019OccupancyNL}, 3D scene surface~\cite{Sitzmann2019SceneRN,Jiang2020LocalIG,Genova2020LocalDI,Sitzmann2020ImplicitNR}, 2D images~\cite{Chen2020LearningCI,Sitzmann2020ImplicitNR} and 1D audios~\cite{Sitzmann2020ImplicitNR}. 
Compared to traditional representations, DIF is shown to be more efficient, expressive, and is fully continuous. 
It is able to capture better structural details with fewer parameters when trained properly.
For example, 
DeepSDF~\cite{Park2019DeepSDFLC} takes a 3D coordinate and a categorical latent code as the input, and outputs the signed distance (SDF) at this coordinate to decide whether it is inside the target shape.
Local Implicit Grid (LIG)~\cite{Jiang2020LocalIG} learns the common geometric features from local overlapping patches and reconstructs complicated scenes by associating them.
Local Implicit Image Function (LIIF)~\cite{Chen2020LearningCI} extracts a set of latent codes distributed in the LR domain to interpolate the HR target image.
SIREN~\cite{Sitzmann2020ImplicitNR} proposed a general implicit neural representation for various domains to fit complicated signals by using periodic activation functions.
Different from previous methods that focus on learning from single-modal data, we stress our work on learning from multi-modal data, e.g., HR RGB guide and LR depth input. 
We focus on extracting prior knowledge from the guide image to help the representation learning of the target image.
On this aspect, our method is more close to PixTransform~\cite{Lutio2019GuidedSA}.
This method explores guided super-resolution from a different perspective more similar to depth estimation, by training a DIF that maps each pixel in the guide image to the target image, and supervises only by the LR input. 
Different from our training pipeline, it doesn't rely on HR target image for supervision, and can be categorized as an unsupervised depth super-resolution task.
Besides, PixTransform is unconditional to observations, which means it needs to train a different set of parameters for every new image. 

\subsection{Graph Attention Mechanism}
Graph Convolution Networks (GCNs) focus on problems residing in graph-structured data, by defining graph convolutions on the vertices and edges of a graph~\cite{Kipf2017SemiSupervisedCW,Velickovic2018GraphAN,Wang2019DynamicGC,tang2021rgln}. 
An undirected graph $\mathcal G=(\mathcal V,\mathcal E)$ is composed of $N$ vertices $v_i \in \mathcal V$, edges $(v_i, v_j) \in \mathcal E$, and an adjacency matrix $A \in \mathbb R^{N \times N}$ which is binary or weighted.
For tasks where $A$ is binary, some work explores to learn the edge weights from vertex features to facilitate the feature propagation.
Graph Attention Networks~\cite{Velickovic2018GraphAN} leverages masked self-attention layers to regress a continuous weight between each two connected vertices. 
EdgeConv~\cite{Wang2019DynamicGC} learns different adjacency matrices in different layers to extract vertex features in a dynamic way. 
For the task of guided super-resolution, we propose to first divide the image into pixels with implicit neural representation, then treat each pixel query as a graph problem.
Thus, the interpolation weights can be interpreted as graph edge weights and learned through the graph attention mechanism.

%% file: method.tex
In this section, we introduce our JIIF representation for guided super-resolution task. 
We first review the recent neural implicit interpolation methods in Section~\ref{sec:nueral_image_interpolation}, 
then detail our JIIF representation for guided super-resolution in Section~\ref{sec:jiif}. 
Finally, we describe our design of the JIIF-Net to learn the representation from data in Section~\ref{sec:network}.

\begin{figure*}[ht]
    \centering
    \includegraphics[width=0.95\textwidth]{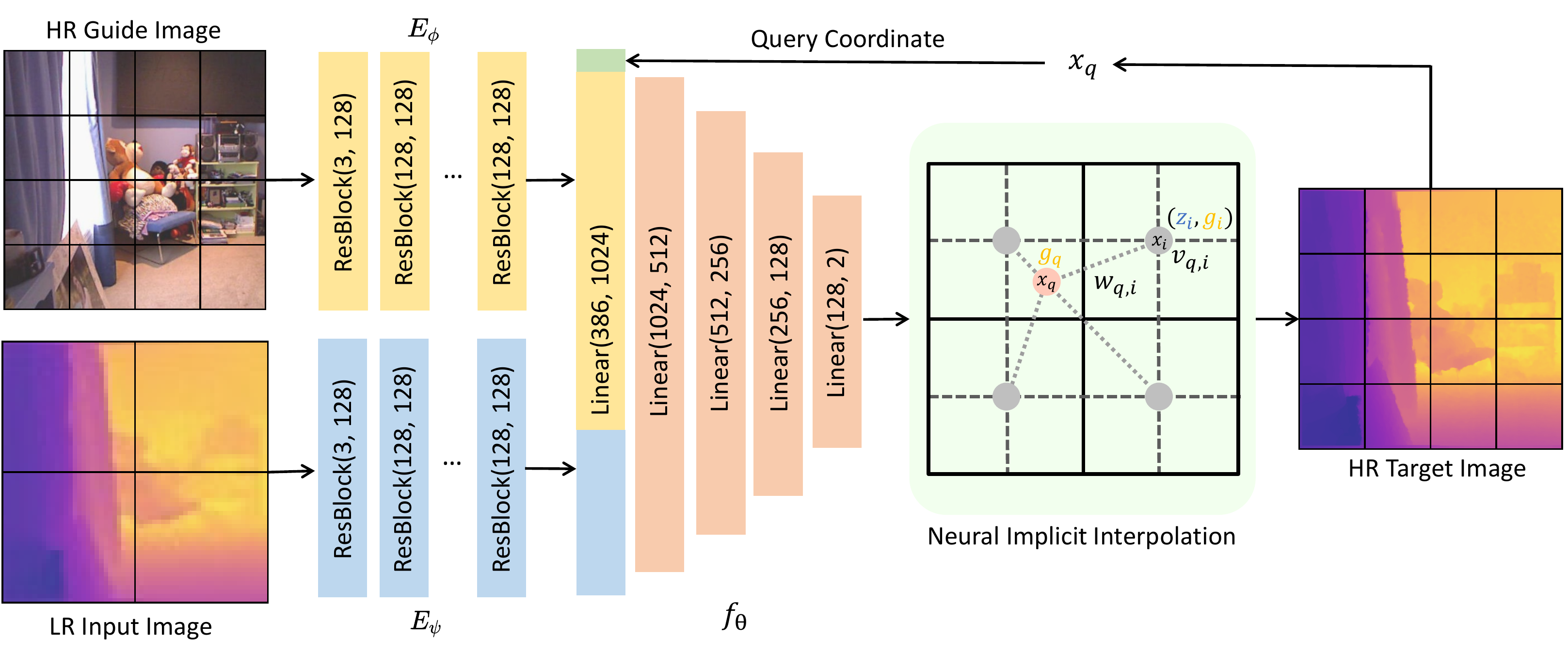}
    \caption{Network Architecture. 
    The grid illustrates the relative image resolution, and we use a $\times2$ up-sampling as an example for simplicity.
    Given a HR guide image and a LR input image, we extract two sets of latent codes via two encoders, then query the JIIF decoder with a coordinate in the HR domain to predict the pixel value at this coordinate. The prediction is a weighted average from the four nearest coordinates in the LR domain just like the standard image interpolation, but we learn the interpolation weights and values via a deep implicit function.
    }
    \label{fig:network}
\end{figure*}

\subsection{Neural Implicit Image Interpolation}
\label{sec:nueral_image_interpolation}

We start from a general formulation of the image interpolation problem for image up-sampling, then view it from the perspective of implicit neural representation to introduce the neural implicit image interpolation method.
For each LR input image $M$, we want to calculate the corresponding HR target image $I$:
\begin{equation}
    I(x_q) = \sum_{i \in \mathcal N_q} w_{q,i} v_{q,i},
\end{equation}
where $x_q$ is the coordinate of the query pixel $q$ in the HR domain, $\mathcal N_q$ is the set of neighbor pixels for $q$ in the LR domain, $w_{q,i}$ is the interpolation weight between $i$ and $q$, and $v_{q,i}$ is the interpolation value for $i$.
The interpolation weights are usually normalized so that $\sum_{i \in \mathcal N_q} w_{q,i} = 1$.
We use a continuous image representation by scaling the image coordinates into $(-1, 1)$ to make it possible to share the coordinate in both the HR and LR domain.
Due to the nature of 2D images, $\mathcal N_q$ is usually chosen as the four nearest corner pixels of $q$ in the LR domain (as illustrated in Figure~\ref{fig:network}).
Different interpolation methods have different ways to calculate the interpolation weights and values.
The most commonly used bilinear interpolation is implemented with:
\begin{equation}
\begin{cases}
    w_{q, i} = \frac {S_i} {S}, \\
    v_{q, i} = M(x_i),
\end{cases}
\end{equation}
where $S_i$ is the partial area diagonally opposite to the corner pixel $i$, $S = \sum_{i \in \mathcal N_q} S_i$ is the total area serving as a normalization factor, and $M(x_i)$ is the pixel value of the LR input image at $x_i$.

In implicit neural representation, instead of directly using the pixel value $M(x_i)$, a DIF is applied to calculate the interpolation value $v_{q,i}$.
For example, LIIF~\cite{Chen2020LearningCI}, LIG~\cite{Jiang2020LocalIG} and SCSSNet~\cite{Rist2020SCSSnetLS,Rist2020SemanticSC} all take the following form:
\begin{equation}
    v_{q, i} = f_\theta(z_i, x_q - x_i),
\end{equation}
where $f_\theta(\cdot)$ is a MLP with parameters $\theta$ that takes a local latent code $z_i$ and a relative coordinate $x_q - x_i$ as input.
In this setting, the target image is represented by a set of local latent codes distributed at the pixel coordinates of the LR domain, 
each storing information about its local area~\cite{Chen2020LearningCI}. 
The latent codes map is the output feature map from an encoder network, and it is of the same resolution as the LR input image:
\begin{equation}
    z_i = E_\phi(M)(x_i),
\end{equation}
where $E_\phi(\cdot)$ is the encoder network with parameters $\phi$.
The DIF thus models a local area centered at the coordinate of the given latent code.
By querying the conditioned DIF $f_\theta(z_i, \cdot)$ with a relative query coordinate $x_q - x_i$, it returns the estimated target value at the query coordinate $x_q$, e.g., the depth value in depth super-resolution.
The weighted average of these estimated values from the four corners is further calculated to avoid discontinuous prediction (which is called local ensemble in~\cite{Chen2020LearningCI}).

\subsection{Joint Implicit Image Function}
\label{sec:jiif}

We focus on the problem of guided super-resolution, where an extra HR guide image $G$ is provided with the LR input image $M$. 
Previous methods either directly regress the target image values by fusing CNN features which lacks interpretability~\cite{li2016deep,Li2019JointIF}, or treat it as an explicit filtering problem which cannot fully take advantage of the information in the guide image~\cite{Kim2021DeformableKN}.
We hypothesize that the information in the guide image can benefit the learning of both interpolation weights and values, and these two terms can be learned jointly to boost the performance. 
Inspired by the recent neural implicit image interpolation methods, we propose to use DIFs to model both the interpolation weights and values, which we call the \textit{Joint Implicit Image Function} representation.

Similar to the LIIF representation, the target image is represented by a set of local latent codes, but our latent codes are extracted from both the LR input image and the HR guide image, allowing the detailed information from the guide image to help the up-sampling process. 
In particular, we apply two encoder networks to extract two sets of latent codes from the guide image and the input image respectively:
\begin{equation}
    \begin{cases}
    z_i = E_\phi(M)(x_i), \\
    g_j = E_\psi(G)(x_j),
    \end{cases}
\label{eq:two-latent-codes}
\end{equation}
where $E_\psi$ is another encoder network with parameters $\psi$.
Then, the interpolation values can be naturally calculated by querying the DIF with these two latent codes and a relative coordinate:
\begin{equation}
    v_{q, i} = f_\theta(z_i, g_i, x_q - x_i),
    \label{eq:jiif_value}
\end{equation}
where $i$ is one of the neighbors of $q$ in the LR domain ($i \in \mathcal N_q $). 
Please note that due to the different resolutions of HR and LR images, we could not obtain the HR latent code at position $x_i$ directly. In such cases, we conduct the bicubic interpolation operation to approximate the HR latent code at position $x_i$.

Furthermore, we propose to learn the interpolation weights at the same time.
As illustrated in the neural implicit interpolation part in Figure~\ref{fig:network}, we view the interpolation at each query pixel as a graph problem.
The four corner pixels and the query pixel are the vertices, and each corner pixel is connected to the query pixel with an edge.
Previous methods usually use an empirical value for the edge weights~\cite{Chen2020LearningCI}, or directly regress the weights from the CNN features~\cite{Kim2021DeformableKN}.
Inspired by recent research in Graph Convolutional Networks~\cite{Velickovic2018GraphAN,Wang2019DynamicGC}, we propose to use a graph attention mechanism to calculate the edge weights.
Specifically, we extract the guide latent code of each corner pixel $g_i$ and the query pixel $g_q$ in the HR domain, and apply a MLP to learn the weight in an asymmetric way:
\begin{equation}
    a_{q,i} = f_\eta(g_i, g_q - g_i),
    \label{eq:jiif_weight}
\end{equation}
where $a_{q,i}$ is the learned edge weight, and $f_\eta$ is a MLP with parameters $\eta$.

We notice the representation of the interpolation weights (Equation~\ref{eq:jiif_weight}) and values (Equation~\ref{eq:jiif_value}) are of a similar form. 
Hence, we propose to integrate these two separate functions into a unified one:
\begin{equation}
    a_{q,i}, v_{q,i} = f_\theta(z_i, g_i, g_q - g_i, x_q - x_i),
    \label{eq:jiif}
\end{equation}

By integrating the learning of interpolation weights and values, we reduce the parameters needed to model the representation, and allow interaction between these two processes, which is demonstrated to be more effective in our experiments.
Finally, the edge weights are normalized by applying the softmax function to calculate the final interpolation weights:
\begin{equation}
    w_{q,i} = \frac {\exp(a_{q,i})} {\sum_{i \in \mathcal N_q}\exp(a_{q,i})},
\end{equation}

\input{tabs/dep_up}

\input{figs/nyu}

\input{tabs/noise_middlebury}
\input{figs/noisy_middlebury}

\subsection{Network Architecture and Training}
\label{sec:network}

After defining the JIIF representation, we design a neural network to learn the representation from large datasets.
As shown in Figure~\ref{fig:network}, the network contains two image encoders and one JIIF decoder.
The input image and the guide image are fed into two encoders respectively, generating two feature maps as the latent codes for the JIIF representation.
During training, we sample a set of pixels from the HR image with their coordinates, and query the JIIF decoder with these coordinates to predict the pixel values.
A standard L1 loss is applied to optimize the network for predicting accurate results:
\begin{equation}
    \mathcal L = \frac 1 N \sum_{i}^{N} | \hat I(x_i) - I(x_i) |,
\end{equation}
where $N$ is the total number of sampled pixels, $x_i$ is the coordinate of any sampled pixel, $I(x_i)$ is the ground truth pixel value, and $\hat I(x_i)$ is the predicted pixel value.
In testing, we query all pixels' coordinates in the target domain to recover the full up-sampled image.

%% file: tabs/dep_up.tex
\begin{table*}[t]
\centering

\caption{Quantitative comparison with the state of the art on depth map upsampling in terms of average RMSE.}
\label{tab:depth-upsampling}

\newcolumntype{L}[1]{>{\raggedright\arraybackslash}p{#1}}
\newcolumntype{C}[1]{>{\centering\arraybackslash}p{#1}}
\newcolumntype{R}[1]{>{\raggedleft\arraybackslash}p{#1}}

\begin{tabular}[c]{L{4cm} C{1.0cm} C{1.0cm} C{1.0cm} C{1.0cm} C{1.0cm} C{1.0cm} C{1.0cm} C{1.0cm} C{1.0cm} C{1.0cm} C{1.0cm} C{1.0cm}} 
\shline

\multicolumn{1}{c}{Datasets} & \multicolumn{3}{c}{Middlebury} & \multicolumn{3}{c}{Lu} & \multicolumn{3}{c}{NYU v2} \\
\cmidrule{1-10}

\multicolumn{1}{c}{Down-sampling Ratio} &  
\multicolumn{1}{c}{$\times4$} &  \multicolumn{1}{c}{$\times8$} &  \multicolumn{1}{c}{$\times16$} &  
\multicolumn{1}{c}{$\times4$} &  \multicolumn{1}{c}{$\times8$} &  \multicolumn{1}{c}{$\times16$} & 
\multicolumn{1}{c}{$\times4$} &  \multicolumn{1}{c}{$\times8$} &  \multicolumn{1}{c}{$\times16$} \\

\cmidrule{1-10}
Bicubic & 2.28 & 3.98 & 6.37 & 2.42 & 4.54 & 7.38 & 4.28 & 7.14 & 11.58 \\
DMSG~\cite{Hui2016DepthMS} & 1.88 & 3.45 & 6.28 & 2.30 & 4.17 & 7.22 & 3.02 & 5.38 & 9.17 \\
{DG}~\cite{Gu2017learning} & {1.97} & {4.16} & {5.27} & {2.06} & {4.19} & {6.90} & {3.68} & {5.78} & {10.08}  \\
{DJF}~\cite{li2016deep} & {1.68} & {3.24} & {5.62} & {1.65} & {3.96} & {6.75} & {2.80} & {5.33} & {9.46}  \\
{DJFR}~\cite{Li2019JointIF} & {1.32} & {3.19} & {5.57} & {1.15} & {3.57} & {6.77} & {2.38} & {4.94} & {9.18}  \\
PAC~\cite{su2019pixel}  & 1.32 & 2.62  & 4.58  & 1.20  & 2.33 & 5.19 & 1.89 &  3.33 & 6.78  \\
DKN~\cite{Kim2021DeformableKN} & 1.23 & {2.12} & {4.24} & 0.96 & 2.16 & 5.11 & {1.62} & {3.26} & {6.51}  \\
\cmidrule{1-10}
Ours & \textbf{1.09} & \textbf{1.82} & \textbf{3.31} & \textbf{0.85} & \textbf{1.73} & \textbf{4.16} & \textbf{1.37} & \textbf{2.76} & \textbf{5.27} \\

\shline

\end{tabular}

\end{table*}

%% file: figs/nyu.tex
\begin{figure*}[t]
  \centering
  \footnotesize	
  \subfloat[{RGB image}]{
    \begin{minipage}[b]{0.16\linewidth}
      \includegraphics[width=\linewidth]{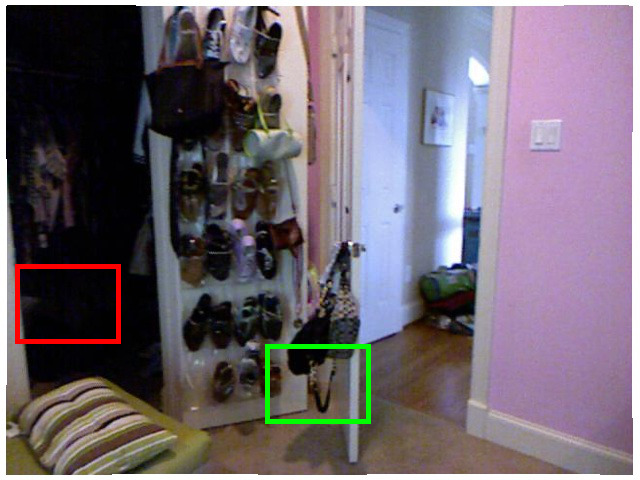}\vspace{-0.05cm}
      \begin{minipage}[b]{0.5\linewidth}\includegraphics[width=\linewidth]{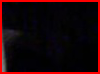}\end{minipage}\hfill
      \begin{minipage}[b]{0.5\linewidth}\includegraphics[width=\linewidth]{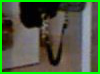}\end{minipage}
      \includegraphics[width=\linewidth]{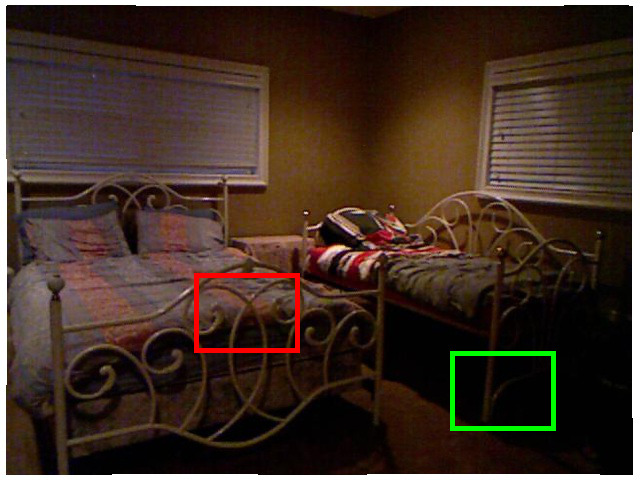}\vspace{-0.05cm}
      \begin{minipage}[b]{0.5\linewidth}\includegraphics[width=\linewidth]{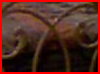}\end{minipage}\hfill
      \begin{minipage}[b]{0.5\linewidth}\includegraphics[width=\linewidth]{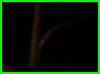}\end{minipage}
      \includegraphics[width=\linewidth]{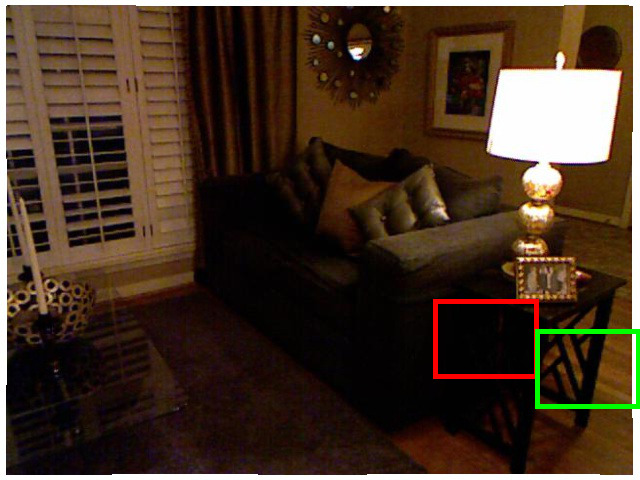}\vspace{-0.05cm}
      \begin{minipage}[b]{0.5\linewidth}\includegraphics[width=\linewidth]{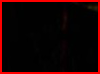}\end{minipage}\hfill
      \begin{minipage}[b]{0.5\linewidth}\includegraphics[width=\linewidth]{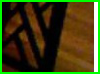}\end{minipage}
      \includegraphics[width=\linewidth]{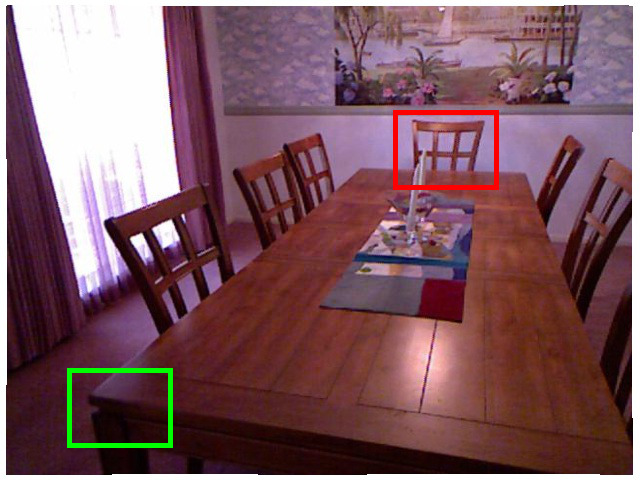}\vspace{-0.05cm}
      \begin{minipage}[b]{0.5\linewidth}\includegraphics[width=\linewidth]{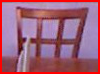}\end{minipage}\hfill
      \begin{minipage}[b]{0.5\linewidth}\includegraphics[width=\linewidth]{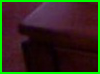}\end{minipage}
      \includegraphics[width=\linewidth]{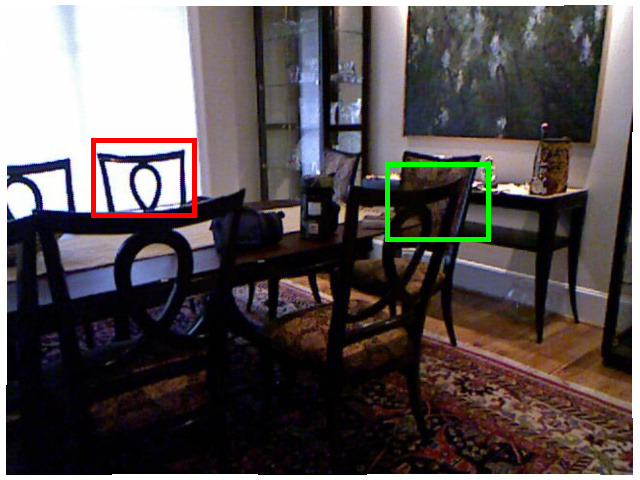}\vspace{-0.05cm}
      \begin{minipage}[b]{0.5\linewidth}\includegraphics[width=\linewidth]{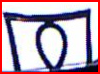}\end{minipage}\hfill
      \begin{minipage}[b]{0.5\linewidth}\includegraphics[width=\linewidth]{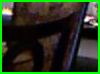}\end{minipage}
    \end{minipage}
  }
  \hspace{-0.12cm}
  \subfloat[{Bicubic Int.}]{
    \begin{minipage}[b]{0.16\linewidth} 
      \includegraphics[width=\linewidth]{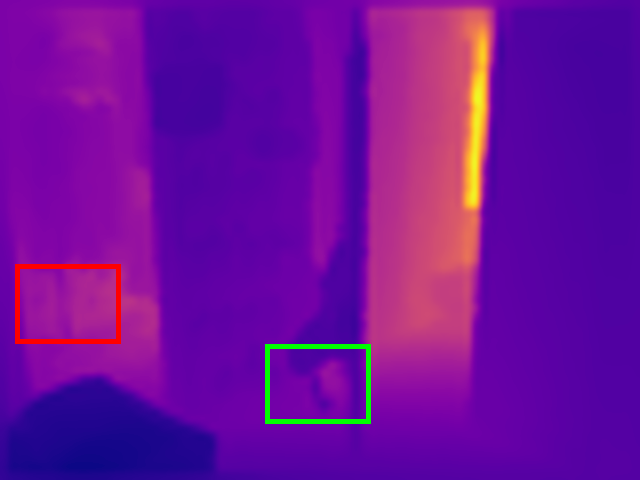}\vspace{-0.05cm}
      \begin{minipage}[b]{0.5\linewidth}\includegraphics[width=\linewidth]{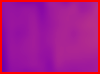}\end{minipage}\hfill
      \begin{minipage}[b]{0.5\linewidth}\includegraphics[width=\linewidth]{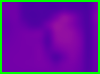}\end{minipage}
      \includegraphics[width=\linewidth]{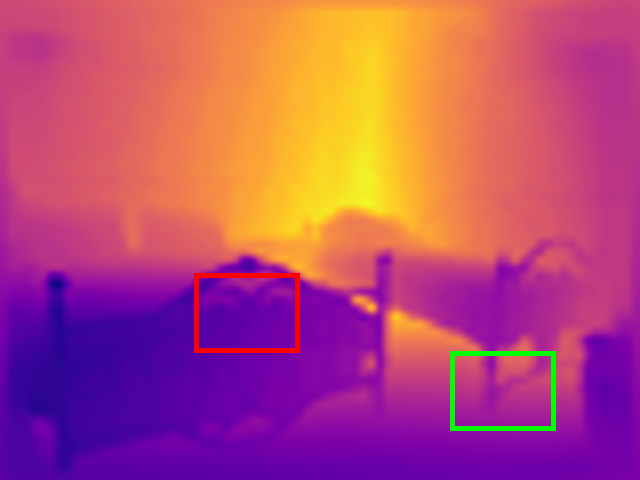}\vspace{-0.05cm}
      \begin{minipage}[b]{0.5\linewidth}\includegraphics[width=\linewidth]{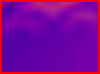}\end{minipage}\hfill
      \begin{minipage}[b]{0.5\linewidth}\includegraphics[width=\linewidth]{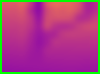}\end{minipage}
      \includegraphics[width=\linewidth]{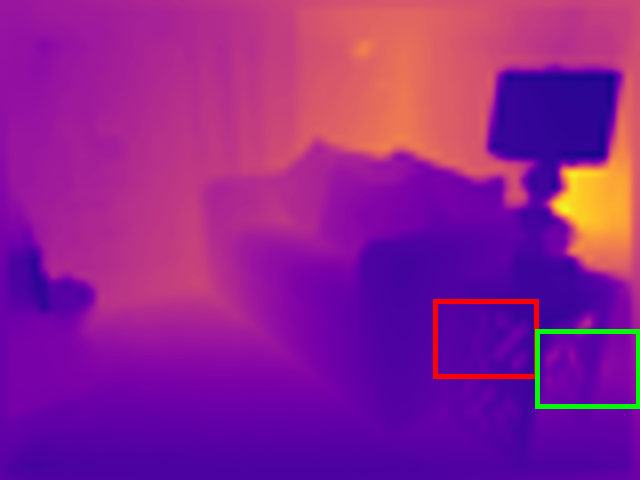}\vspace{-0.05cm}
      \begin{minipage}[b]{0.5\linewidth}\includegraphics[width=\linewidth]{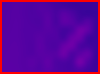}\end{minipage}\hfill
      \begin{minipage}[b]{0.5\linewidth}\includegraphics[width=\linewidth]{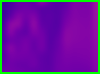}\end{minipage}
      \includegraphics[width=\linewidth]{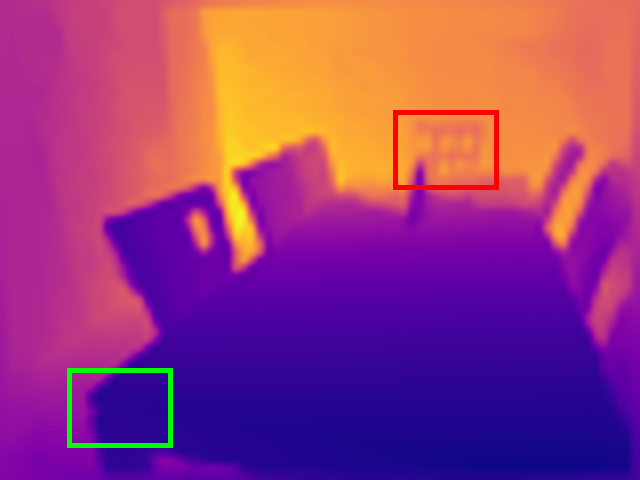}\vspace{-0.05cm}
      \begin{minipage}[b]{0.5\linewidth}\includegraphics[width=\linewidth]{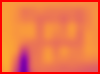}\end{minipage}\hfill
      \begin{minipage}[b]{0.5\linewidth}\includegraphics[width=\linewidth]{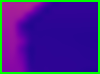}\end{minipage}
      \includegraphics[width=\linewidth]{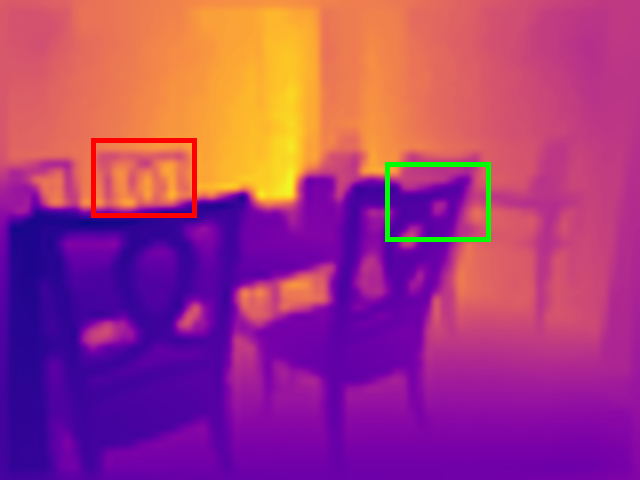}\vspace{-0.05cm}
      \begin{minipage}[b]{0.5\linewidth}\includegraphics[width=\linewidth]{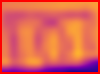}\end{minipage}\hfill
      \begin{minipage}[b]{0.5\linewidth}\includegraphics[width=\linewidth]{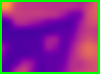}\end{minipage}
    \end{minipage}    
  }
  \hspace{-0.12cm}
  \subfloat[{DJFR~\cite{Li2019JointIF}}]{
    \begin{minipage}[b]{0.16\linewidth} 
      \includegraphics[width=\linewidth]{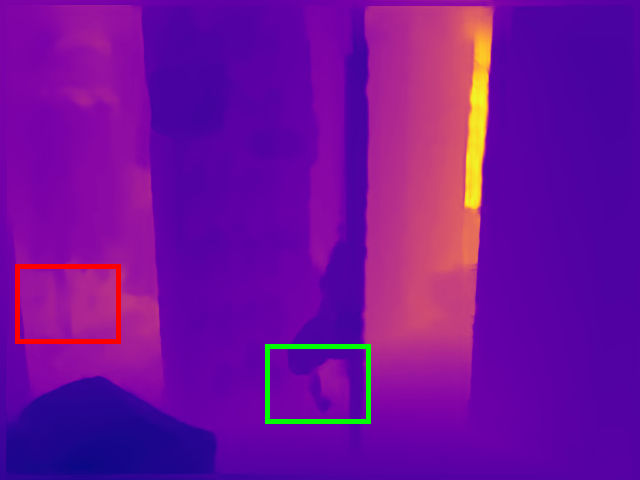}\vspace{-0.05cm}
      \begin{minipage}[b]{0.5\linewidth}\includegraphics[width=\linewidth]{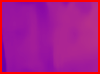}\end{minipage}\hfill
      \begin{minipage}[b]{0.5\linewidth}\includegraphics[width=\linewidth]{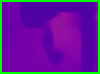}\end{minipage}
      \includegraphics[width=\linewidth]{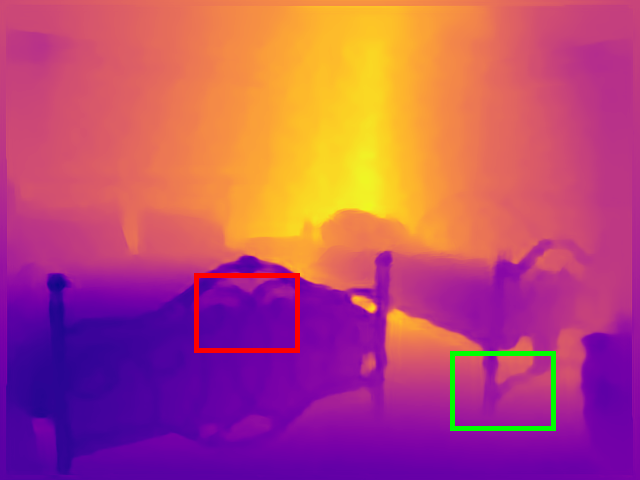}\vspace{-0.05cm}
      \begin{minipage}[b]{0.5\linewidth}\includegraphics[width=\linewidth]{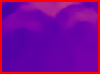}\end{minipage}\hfill
      \begin{minipage}[b]{0.5\linewidth}\includegraphics[width=\linewidth]{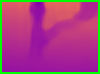}\end{minipage}
      \includegraphics[width=\linewidth]{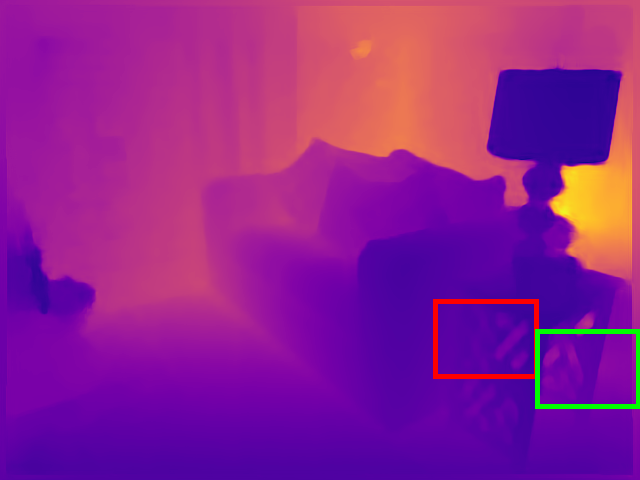}\vspace{-0.05cm}
      \begin{minipage}[b]{0.5\linewidth}\includegraphics[width=\linewidth]{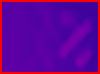}\end{minipage}\hfill
      \begin{minipage}[b]{0.5\linewidth}\includegraphics[width=\linewidth]{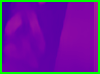}\end{minipage}
      \includegraphics[width=\linewidth]{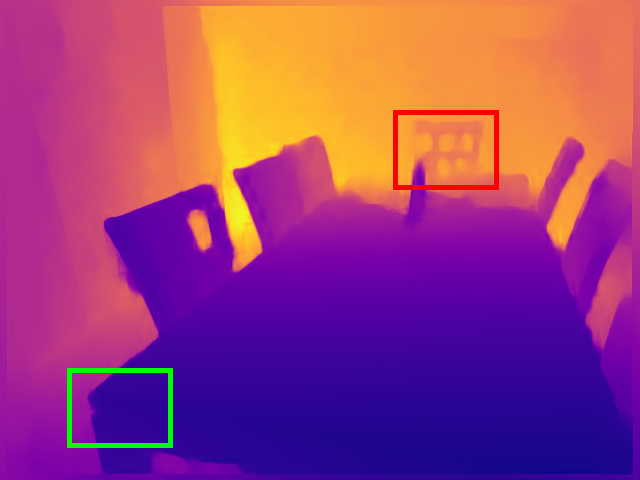}\vspace{-0.05cm}
      \begin{minipage}[b]{0.5\linewidth}\includegraphics[width=\linewidth]{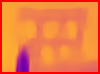}\end{minipage}\hfill
      \begin{minipage}[b]{0.5\linewidth}\includegraphics[width=\linewidth]{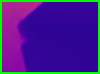}\end{minipage}
      \includegraphics[width=\linewidth]{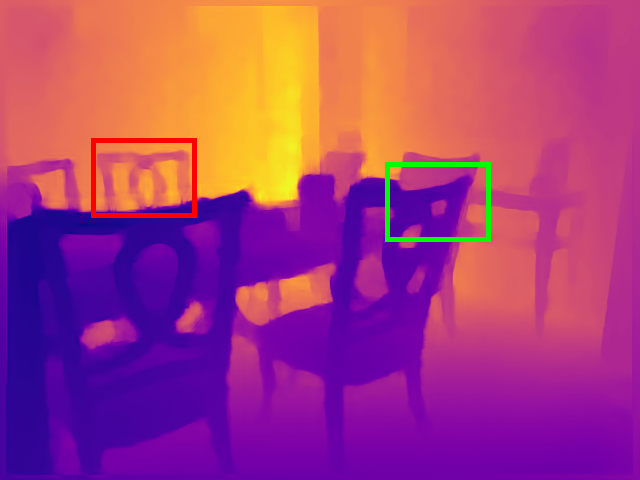}\vspace{-0.05cm}
      \begin{minipage}[b]{0.5\linewidth}\includegraphics[width=\linewidth]{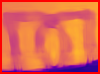}\end{minipage}\hfill
      \begin{minipage}[b]{0.5\linewidth}\includegraphics[width=\linewidth]{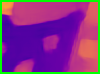}\end{minipage}
    \end{minipage}    
  }
  \hspace{-0.12cm}
  \subfloat[{DKN~\cite{Kim2021DeformableKN}}]{
    \begin{minipage}[b]{0.16\linewidth} 
      \includegraphics[width=\linewidth]{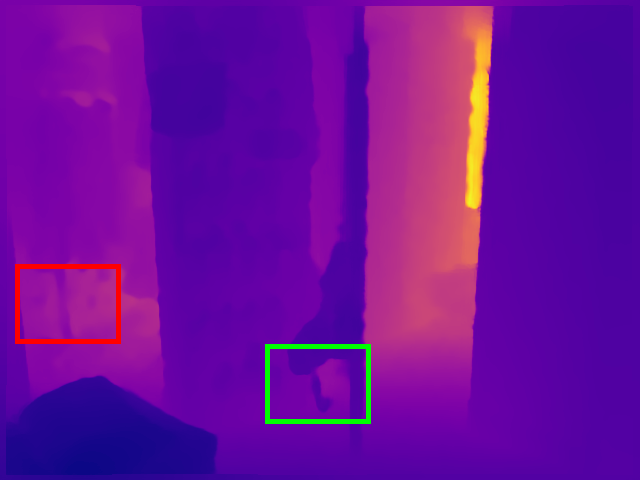}\vspace{-0.05cm}
      \begin{minipage}[b]{0.5\linewidth}\includegraphics[width=\linewidth]{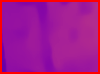}\end{minipage}\hfill
      \begin{minipage}[b]{0.5\linewidth}\includegraphics[width=\linewidth]{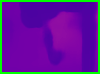}\end{minipage}
      \includegraphics[width=\linewidth]{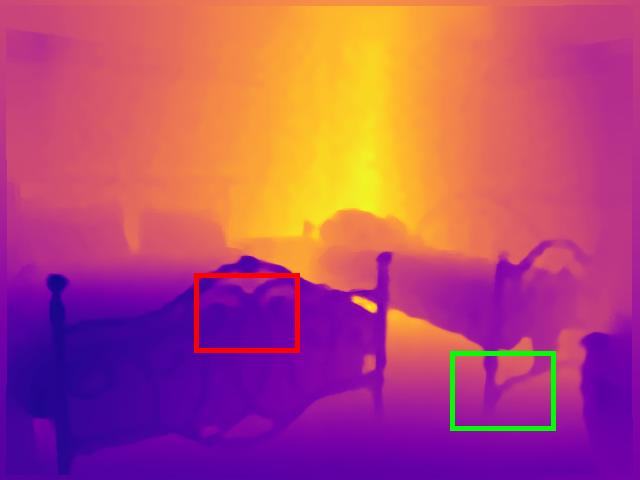}\vspace{-0.05cm}
      \begin{minipage}[b]{0.5\linewidth}\includegraphics[width=\linewidth]{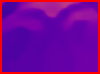}\end{minipage}\hfill
      \begin{minipage}[b]{0.5\linewidth}\includegraphics[width=\linewidth]{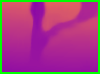}\end{minipage}
      \includegraphics[width=\linewidth]{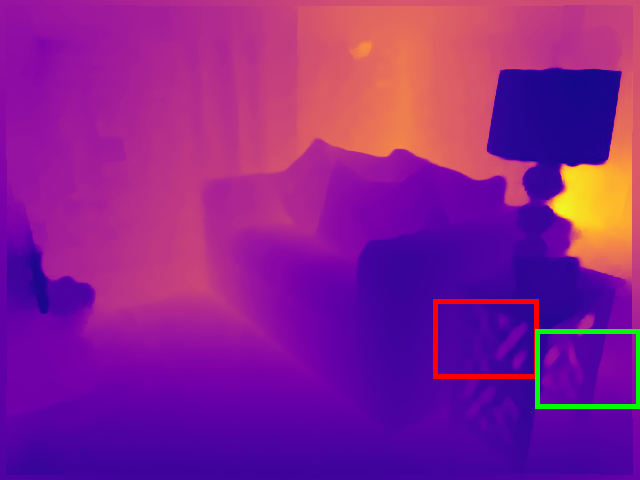}\vspace{-0.05cm}
      \begin{minipage}[b]{0.5\linewidth}\includegraphics[width=\linewidth]{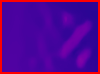}\end{minipage}\hfill
      \begin{minipage}[b]{0.5\linewidth}\includegraphics[width=\linewidth]{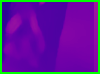}\end{minipage}
      \includegraphics[width=\linewidth]{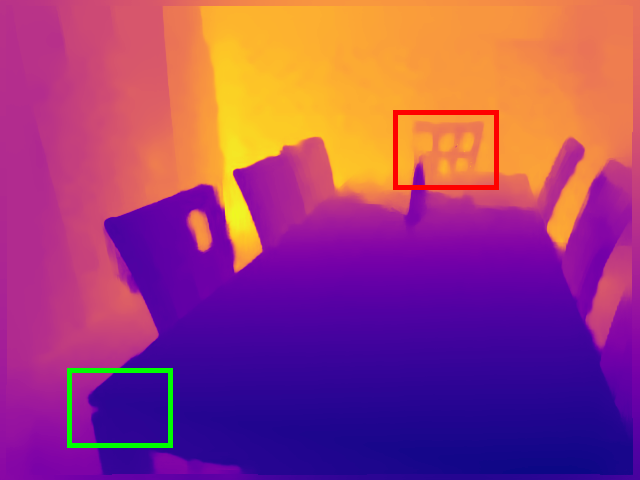}\vspace{-0.05cm}
      \begin{minipage}[b]{0.5\linewidth}\includegraphics[width=\linewidth]{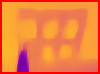}\end{minipage}\hfill
      \begin{minipage}[b]{0.5\linewidth}\includegraphics[width=\linewidth]{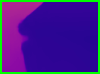}\end{minipage}
      \includegraphics[width=\linewidth]{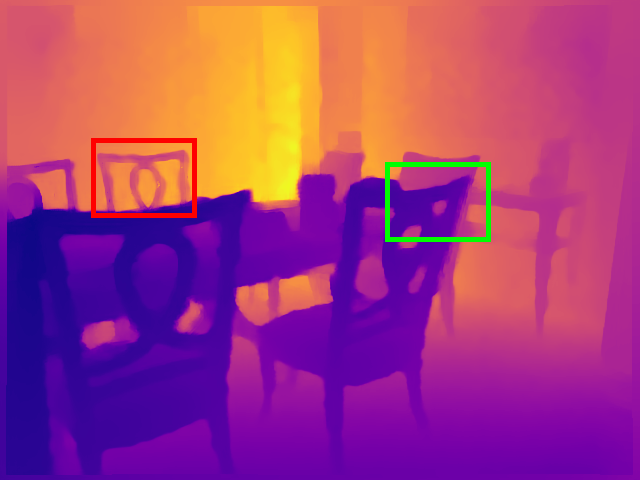}\vspace{-0.05cm}
      \begin{minipage}[b]{0.5\linewidth}\includegraphics[width=\linewidth]{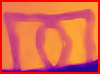}\end{minipage}\hfill
      \begin{minipage}[b]{0.5\linewidth}\includegraphics[width=\linewidth]{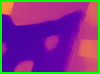}\end{minipage}
    \end{minipage}    
  }
  \hspace{-0.12cm}
  \subfloat[{Ours}]{
    \begin{minipage}[b]{0.16\linewidth} 
      \includegraphics[width=\linewidth]{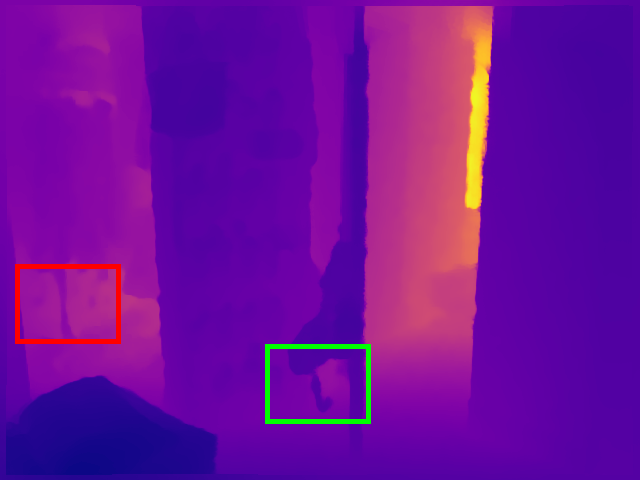}\vspace{-0.05cm}
      \begin{minipage}[b]{0.5\linewidth}\includegraphics[width=\linewidth]{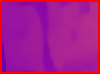}\end{minipage}\hfill
      \begin{minipage}[b]{0.5\linewidth}\includegraphics[width=\linewidth]{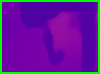}\end{minipage}
      \includegraphics[width=\linewidth]{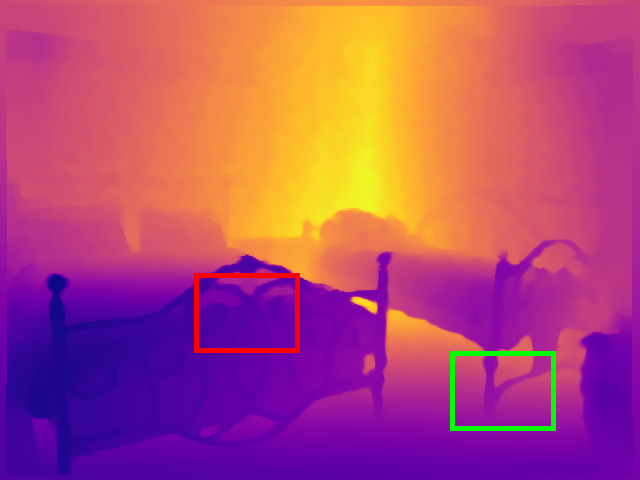}\vspace{-0.05cm}
      \begin{minipage}[b]{0.5\linewidth}\includegraphics[width=\linewidth]{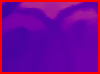}\end{minipage}\hfill
      \begin{minipage}[b]{0.5\linewidth}\includegraphics[width=\linewidth]{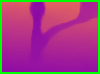}\end{minipage}
      \includegraphics[width=\linewidth]{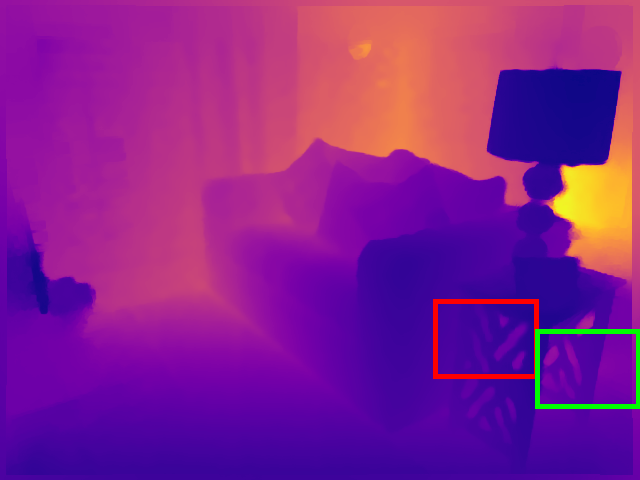}\vspace{-0.05cm}
      \begin{minipage}[b]{0.5\linewidth}\includegraphics[width=\linewidth]{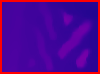}\end{minipage}\hfill
      \begin{minipage}[b]{0.5\linewidth}\includegraphics[width=\linewidth]{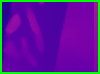}\end{minipage}
      \includegraphics[width=\linewidth]{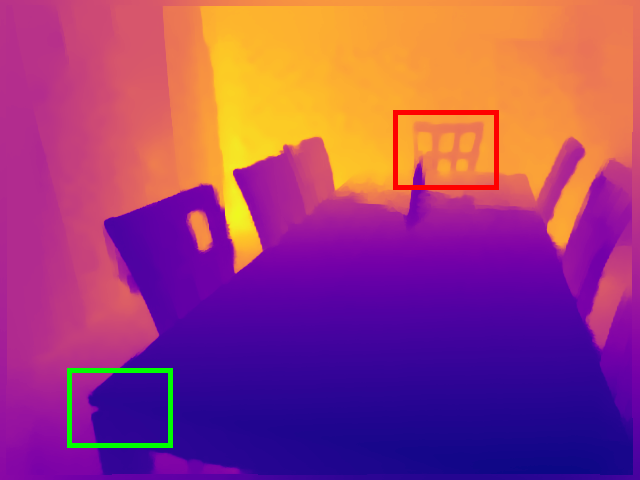}\vspace{-0.05cm}
      \begin{minipage}[b]{0.5\linewidth}\includegraphics[width=\linewidth]{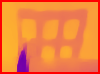}\end{minipage}\hfill
      \begin{minipage}[b]{0.5\linewidth}\includegraphics[width=\linewidth]{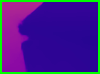}\end{minipage}
      \includegraphics[width=\linewidth]{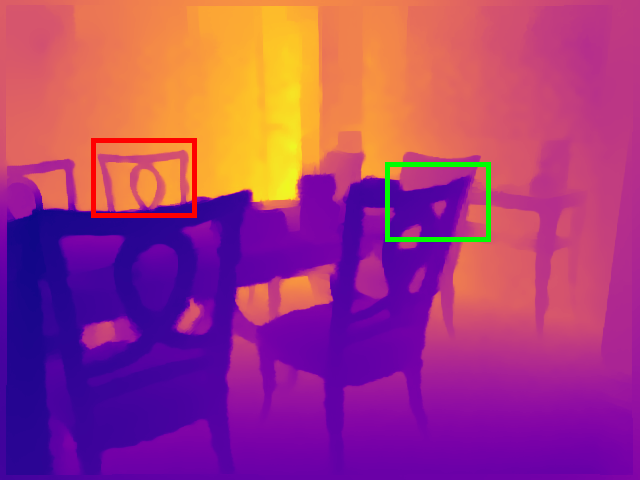}\vspace{-0.05cm}
      \begin{minipage}[b]{0.5\linewidth}\includegraphics[width=\linewidth]{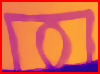}\end{minipage}\hfill
      \begin{minipage}[b]{0.5\linewidth}\includegraphics[width=\linewidth]{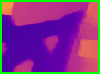}\end{minipage}
    \end{minipage}    
  }
  \hspace{-0.12cm}
  \subfloat[{Ground truth}]{
    \begin{minipage}[b]{0.16\linewidth} 
      \includegraphics[width=\linewidth]{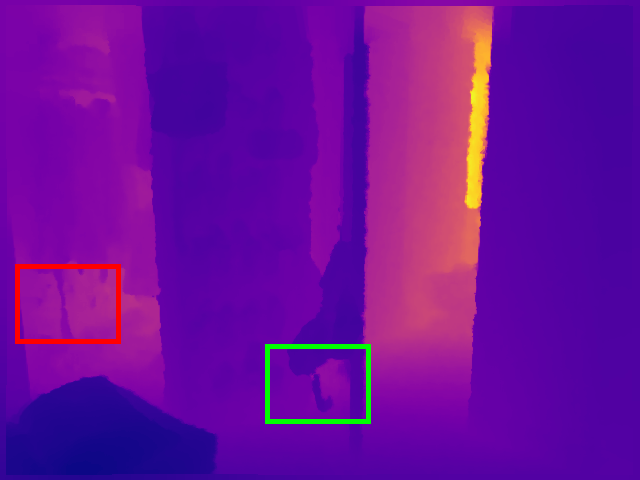}\vspace{-0.05cm}
      \begin{minipage}[b]{0.5\linewidth}\includegraphics[width=\linewidth]{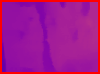}\end{minipage}\hfill
      \begin{minipage}[b]{0.5\linewidth}\includegraphics[width=\linewidth]{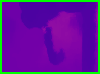}\end{minipage}
      \includegraphics[width=\linewidth]{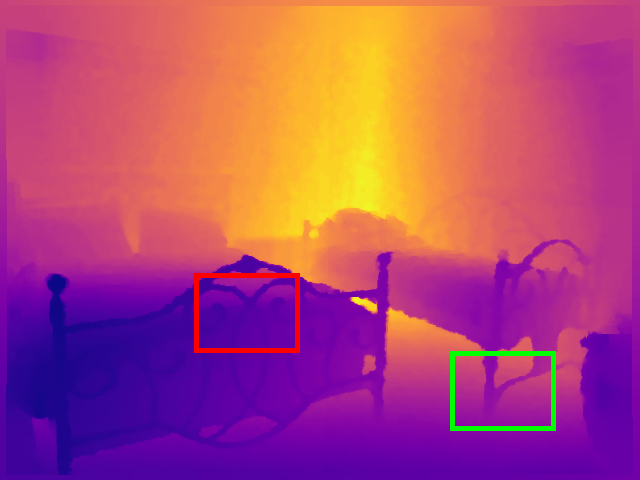}\vspace{-0.05cm}
      \begin{minipage}[b]{0.5\linewidth}\includegraphics[width=\linewidth]{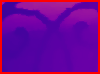}\end{minipage}\hfill
      \begin{minipage}[b]{0.5\linewidth}\includegraphics[width=\linewidth]{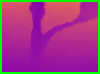}\end{minipage}
      \includegraphics[width=\linewidth]{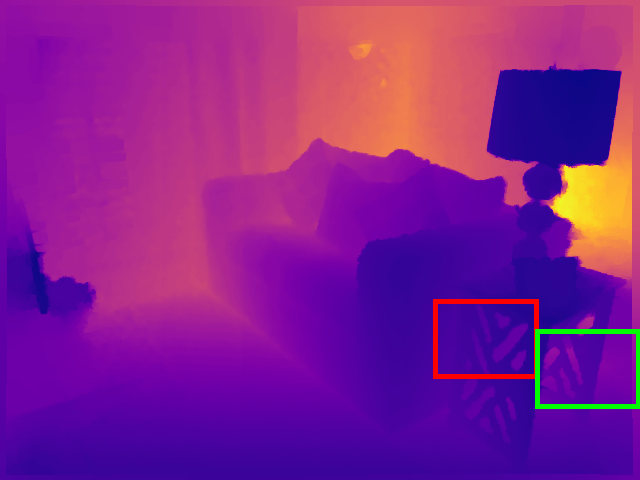}\vspace{-0.05cm}
      \begin{minipage}[b]{0.5\linewidth}\includegraphics[width=\linewidth]{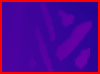}\end{minipage}\hfill
      \begin{minipage}[b]{0.5\linewidth}\includegraphics[width=\linewidth]{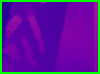}\end{minipage}
      \includegraphics[width=\linewidth]{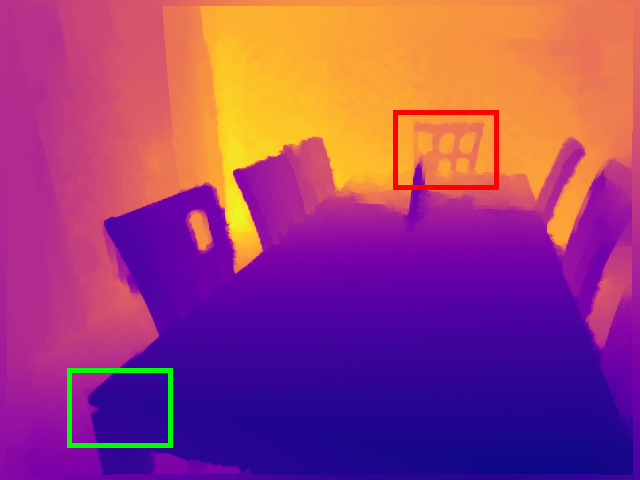}\vspace{-0.05cm}
      \begin{minipage}[b]{0.5\linewidth}\includegraphics[width=\linewidth]{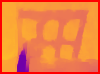}\end{minipage}\hfill
      \begin{minipage}[b]{0.5\linewidth}\includegraphics[width=\linewidth]{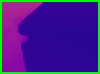}\end{minipage}
      \includegraphics[width=\linewidth]{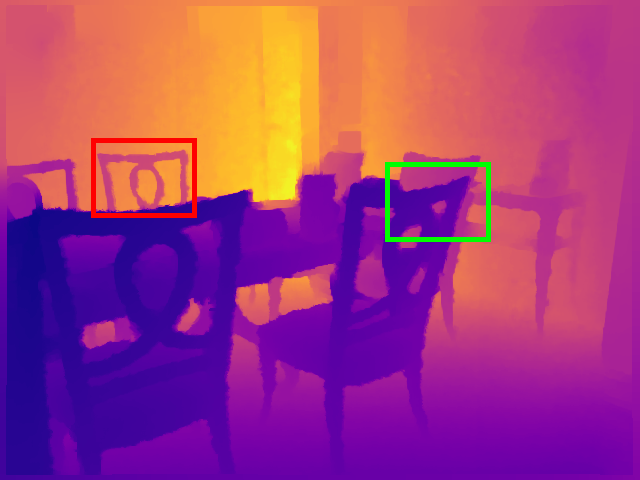}\vspace{-0.05cm}
      \begin{minipage}[b]{0.5\linewidth}\includegraphics[width=\linewidth]{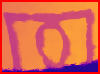}\end{minipage}\hfill
      \begin{minipage}[b]{0.5\linewidth}\includegraphics[width=\linewidth]{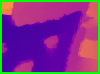}\end{minipage}
    \end{minipage}    
  }
  \caption{Qualitative comparisons of $\times8$ guided depth map super-resolution on the NYU v2 dataset.}
  \label{fig:depth_upsampling}
\end{figure*}

%% file: tabs/noise_middlebury.tex
\begin{table*}[t]
\centering

\caption{Quantitative comparison with the state of the art on noisy depth map upsampling in terms of average RMSE.}
\label{tab:noise-middlebury}
\newcolumntype{L}[1]{>{\raggedright\arraybackslash}p{#1}}
\newcolumntype{C}[1]{>{\centering\arraybackslash}p{#1}}
\newcolumntype{R}[1]{>{\raggedleft\arraybackslash}p{#1}}

\begin{tabular}[c]{L{4cm} C{1.0cm} C{1.0cm} C{1.0cm} C{1.0cm} C{1.0cm} C{1.0cm} C{1.0cm} C{1.0cm} C{1.0cm} C{1.0cm} C{1.0cm} C{1.0cm}} 

\shline

\multicolumn{1}{c}{Datasets} & \multicolumn{3}{c}{Art} & \multicolumn{3}{c}{Books} & \multicolumn{3}{c}{Moebius} \\
\cmidrule(lr){1-10}
\multicolumn{1}{c}{Down-sampling Ratio} & 
\multicolumn{1}{c}{$\times4$} &  \multicolumn{1}{c}{$\times8$} &  \multicolumn{1}{c}{$\times16$} &
\multicolumn{1}{c}{$\times4$} &  \multicolumn{1}{c}{$\times8$} &  \multicolumn{1}{c}{$\times16$} & 
\multicolumn{1}{c}{$\times4$} &  \multicolumn{1}{c}{$\times8$} &  \multicolumn{1}{c}{$\times16$} \\

\cmidrule{1-10}

Bicubic  & 6.07 & 7.27 & 9.59  & 5.15 & 5.45 & 5.97  & 5.51 & 5.68 & 6.11 \\
DMSG~\cite{Hui2016DepthMS}  & 6.19 & 7.26 & 9.53 & 5.38 & 5.18 & 5.20 & 5.48 & 5.06 & 5.36 \\
PDN~\cite{riegler16gdsr} & 3.11 & 4.48 & {7.35} & 1.56 & 2.24 & 3.46 & 1.68 & 2.48 & {3.62} \\
DG~\citep{Gu2017learning} & {2.96} & 4.41 & {7.06} & 1.64 & 2.35 & 3.50 & 1.74 & 2.57 & 3.79 \\
DJFR~\cite{Li2019JointIF} & 4.25 & 6.43 & 9.05  & 2.20 & 3.35 & 4.94 & 2.39 & 3.51 & 4.56 \\ 
PAC~\cite{su2019pixel}  & 5.34 & 7.69 & 10.66 & 2.11 & 3.12 & 4.60 & 2.21 & 3.38 & 4.72 \\
DKN~\cite{Kim2021DeformableKN}  & {3.01} & {4.14} & \textbf{7.01} & {1.44} & {2.10} & {3.09} & {1.63} & {2.39} & {3.55} \\ 

\cmidrule{1-10}

Ours & \textbf{2.79} & \textbf{3.87} & 7.14 & \textbf{1.30} & \textbf{1.75} & \textbf{2.47} & \textbf{1.40} & \textbf{2.03} & \textbf{3.18} \\

\shline

\end{tabular}

\end{table*}

%% file: figs/noisy_middlebury.tex
\begin{figure*}[t]
  \centering
  \footnotesize	
  \subfloat[{RGB image}]{
    \begin{minipage}[b]{0.19\linewidth}
      \includegraphics[width=\linewidth]{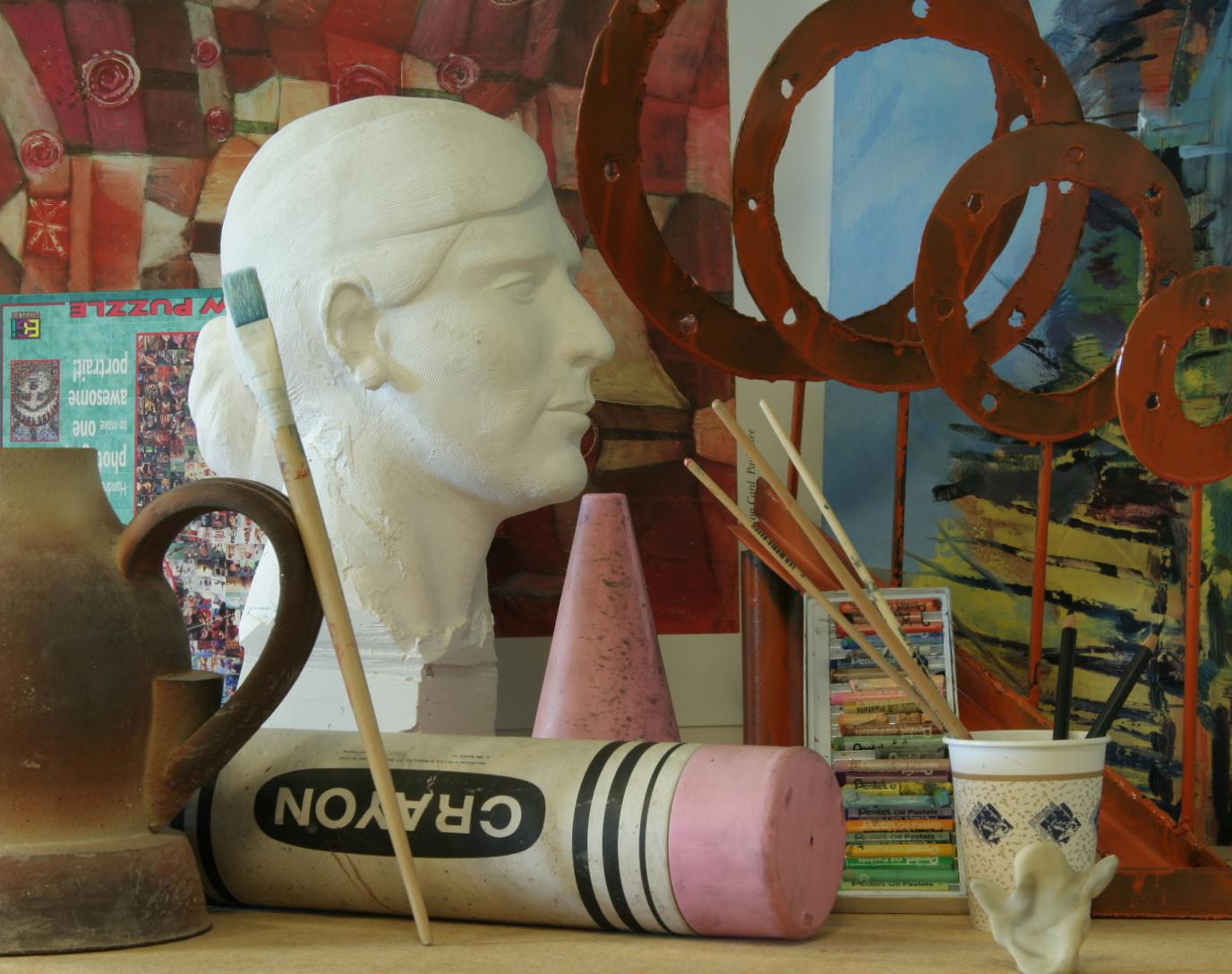}\vspace{-0.05cm}
      \includegraphics[width=\linewidth]{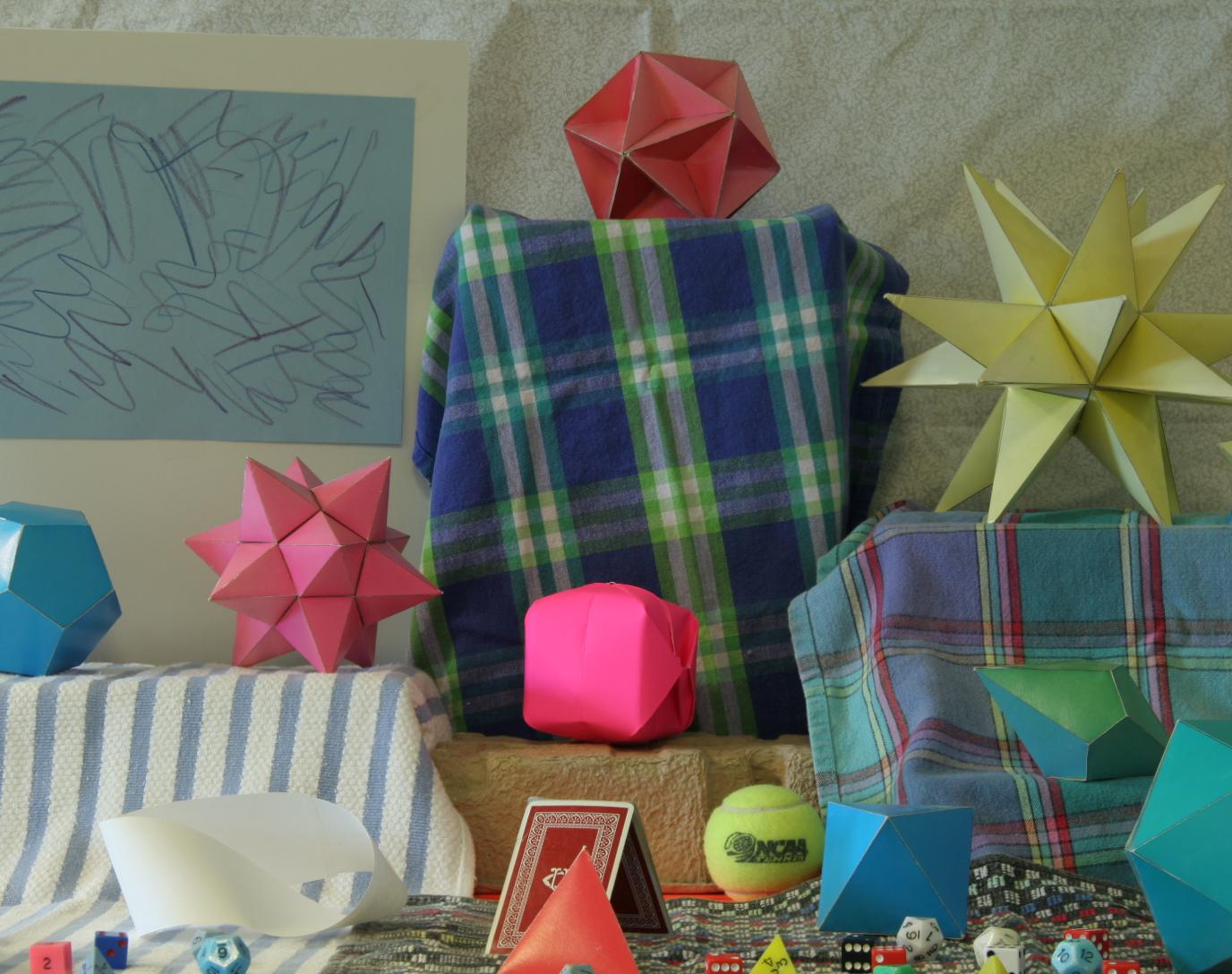}\vspace{-0.05cm}
    \end{minipage}
  }
  \hspace{-0.15cm}
  \subfloat[{Bicubic Int.}]{
    \begin{minipage}[b]{0.19\linewidth} 
      \includegraphics[width=\linewidth]{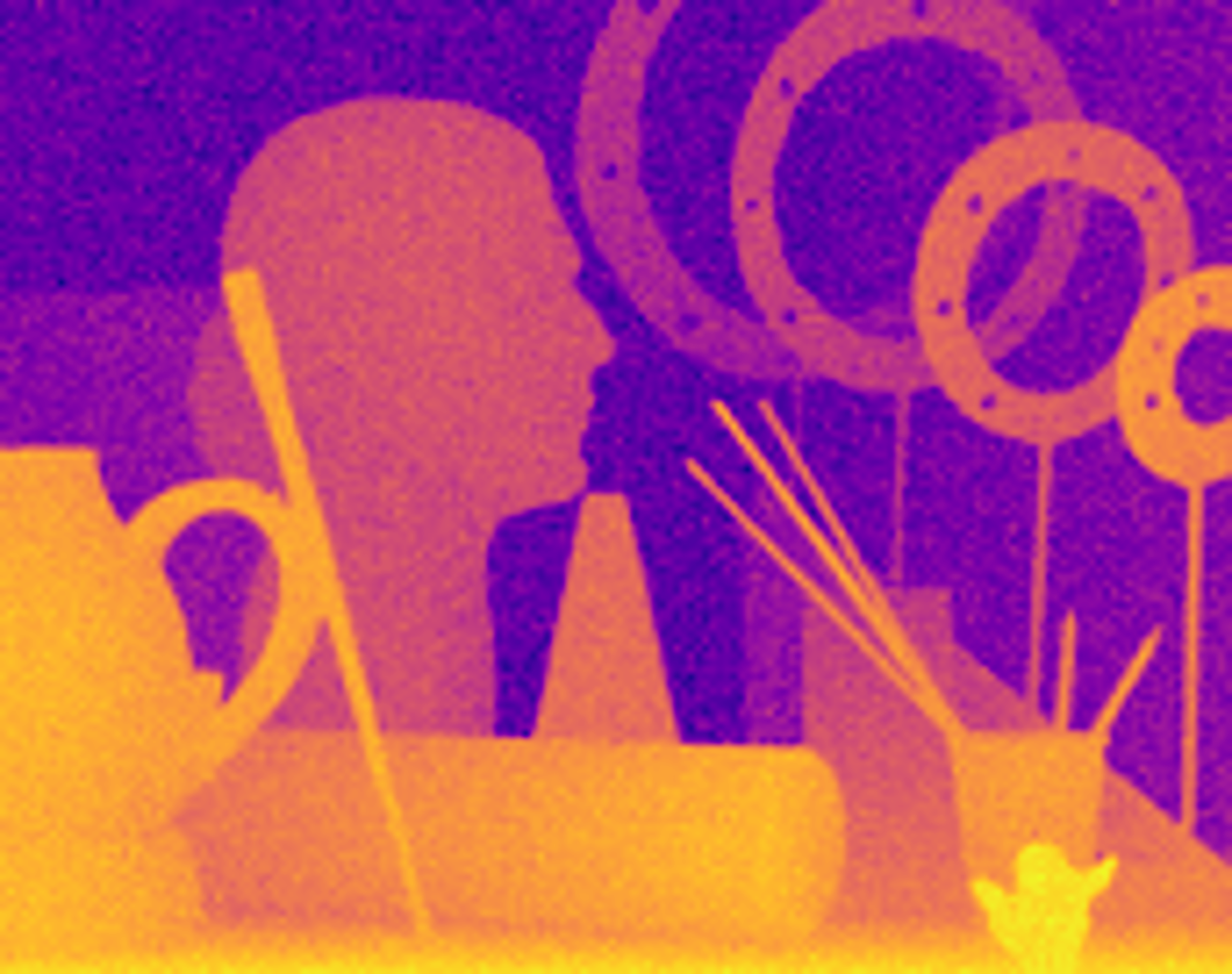}\vspace{-0.05cm}
      \includegraphics[width=\linewidth]{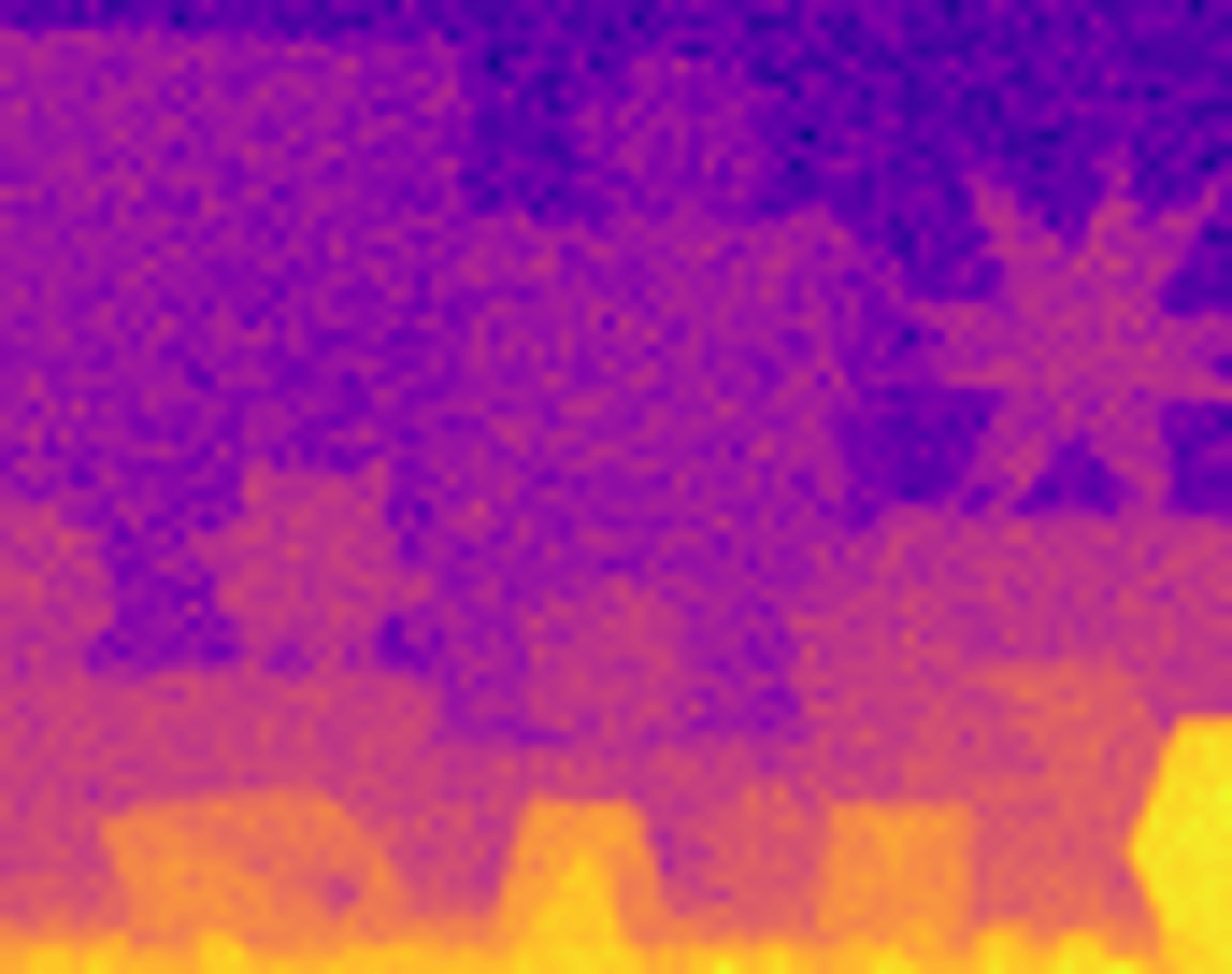}\vspace{-0.05cm}
    \end{minipage}    
  }
  \hspace{-0.15cm}
  \subfloat[{DJFR~\cite{Li2019JointIF}}]{
    \begin{minipage}[b]{0.19\linewidth} 
      \includegraphics[width=\linewidth]{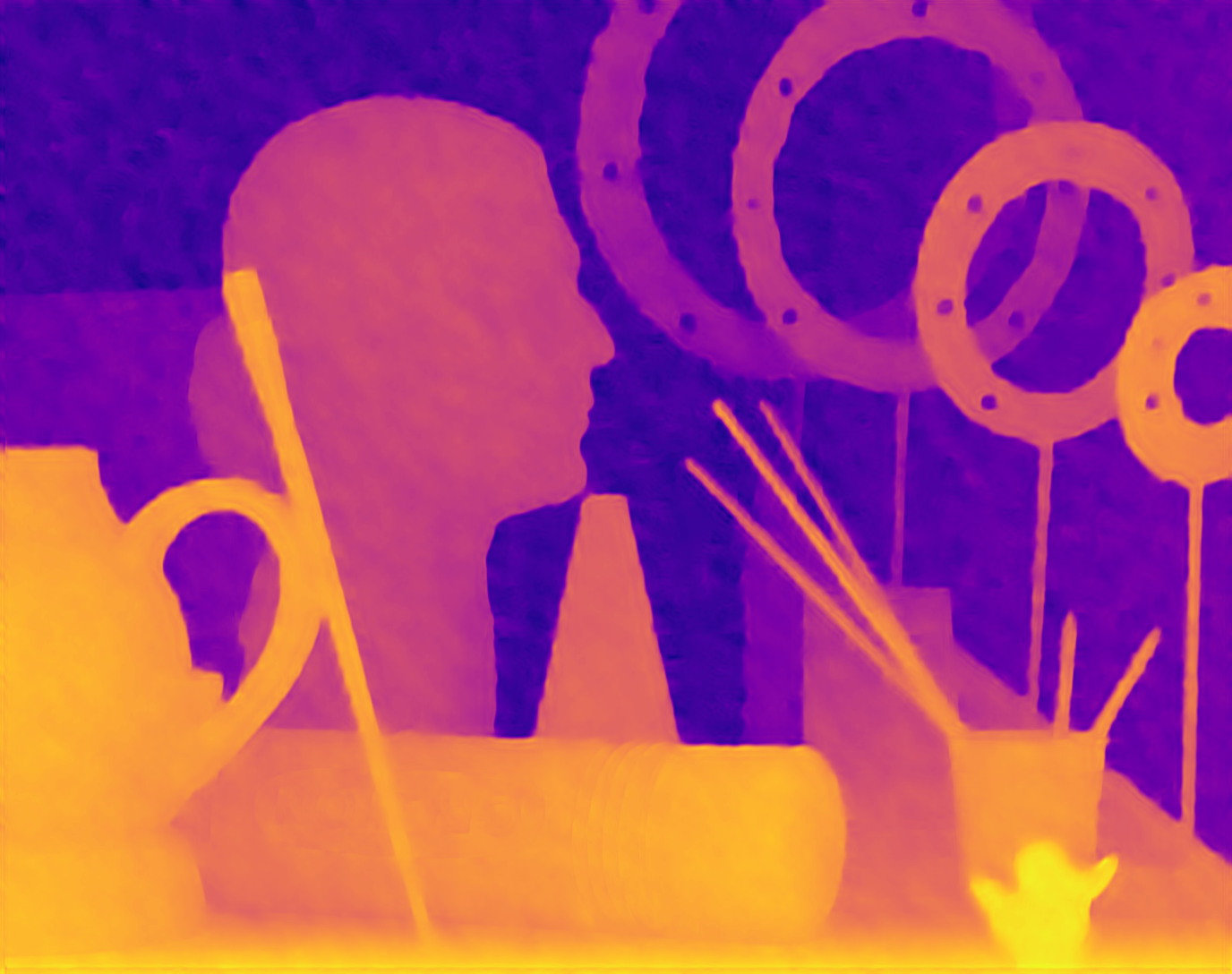}\vspace{-0.05cm}
      \includegraphics[width=\linewidth]{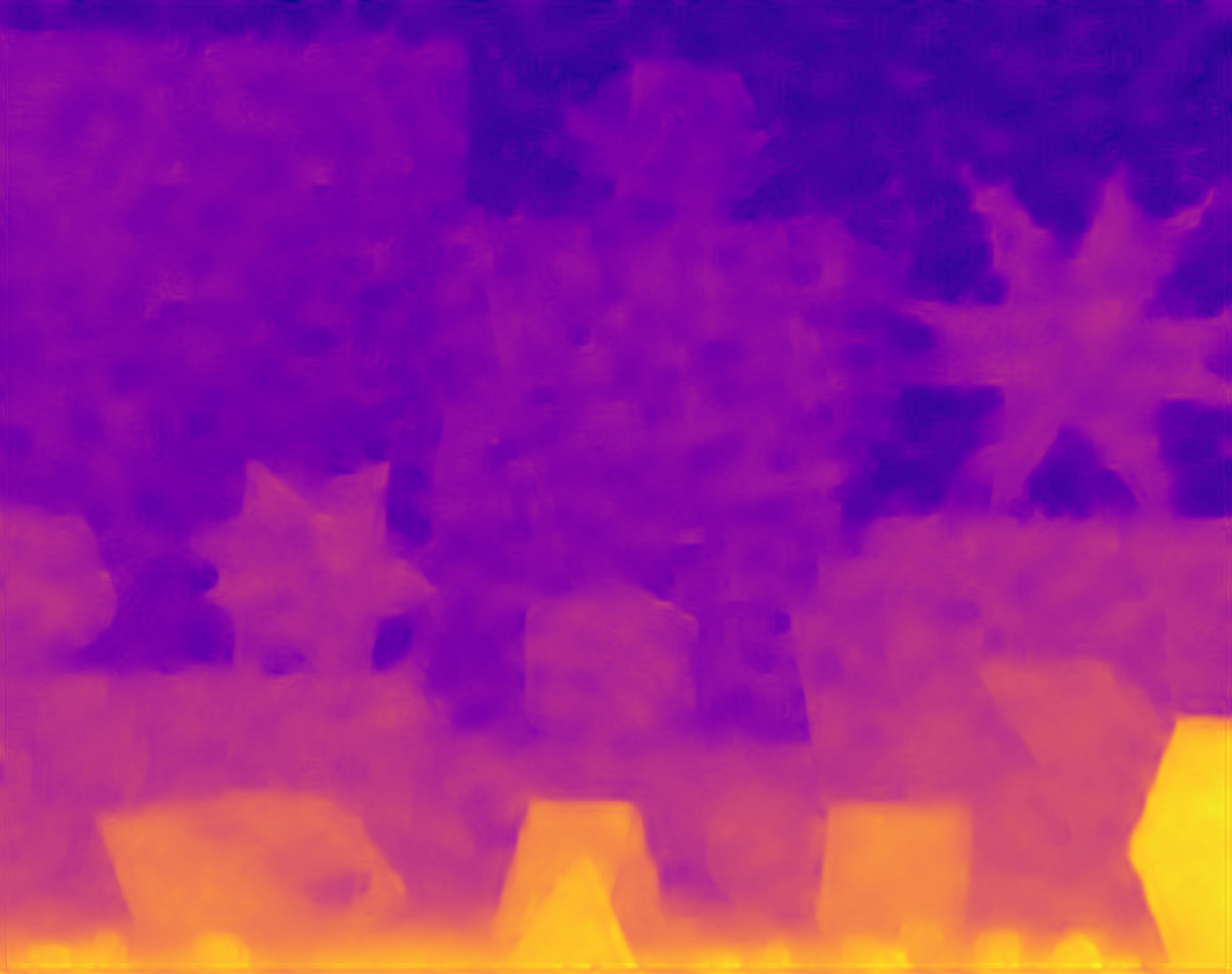}\vspace{-0.05cm}
    \end{minipage}    
  }
  \hspace{-0.15cm}
  \subfloat[{Ours}]{
    \begin{minipage}[b]{0.19\linewidth} 
      \includegraphics[width=\linewidth]{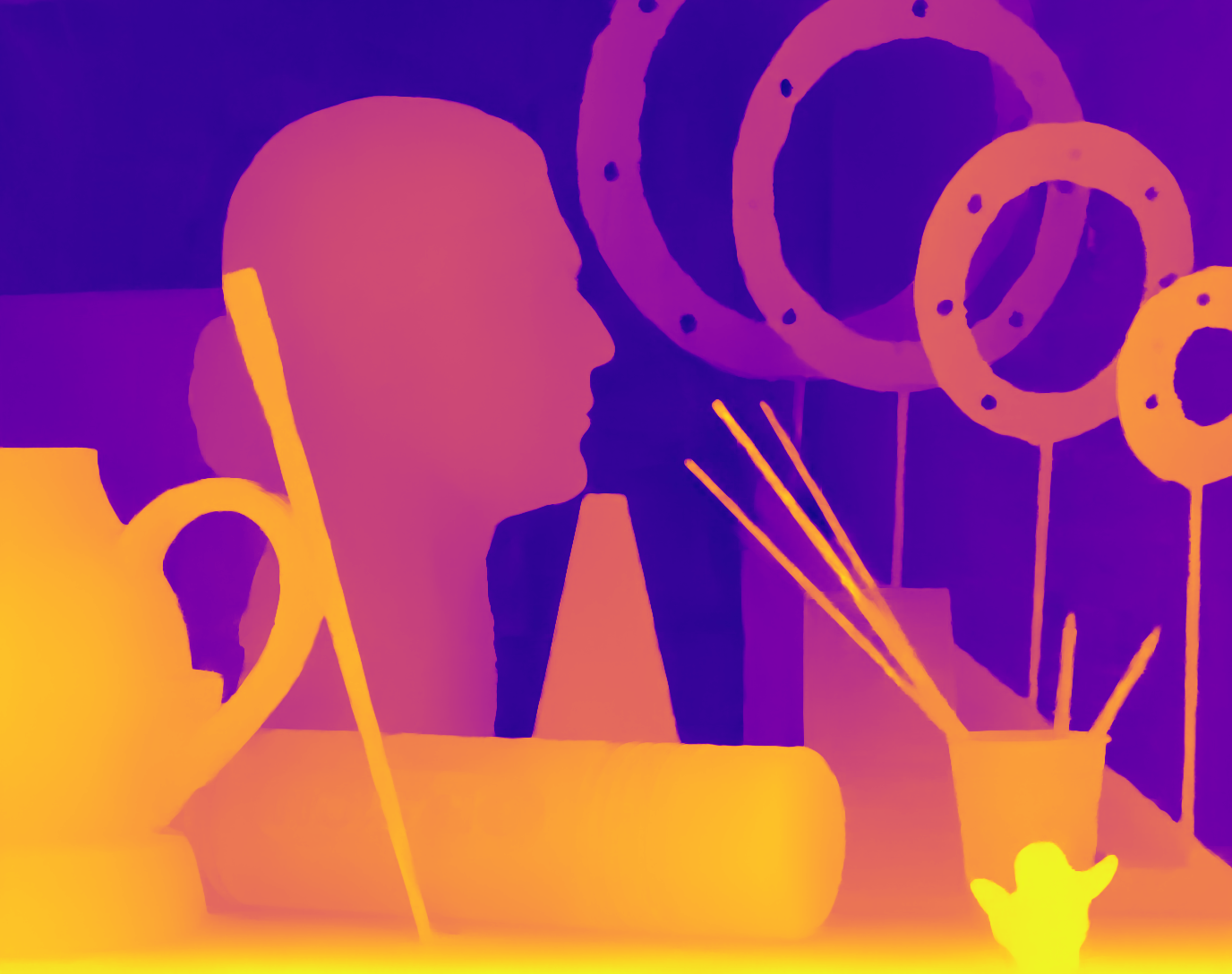}\vspace{-0.05cm}
      \includegraphics[width=\linewidth]{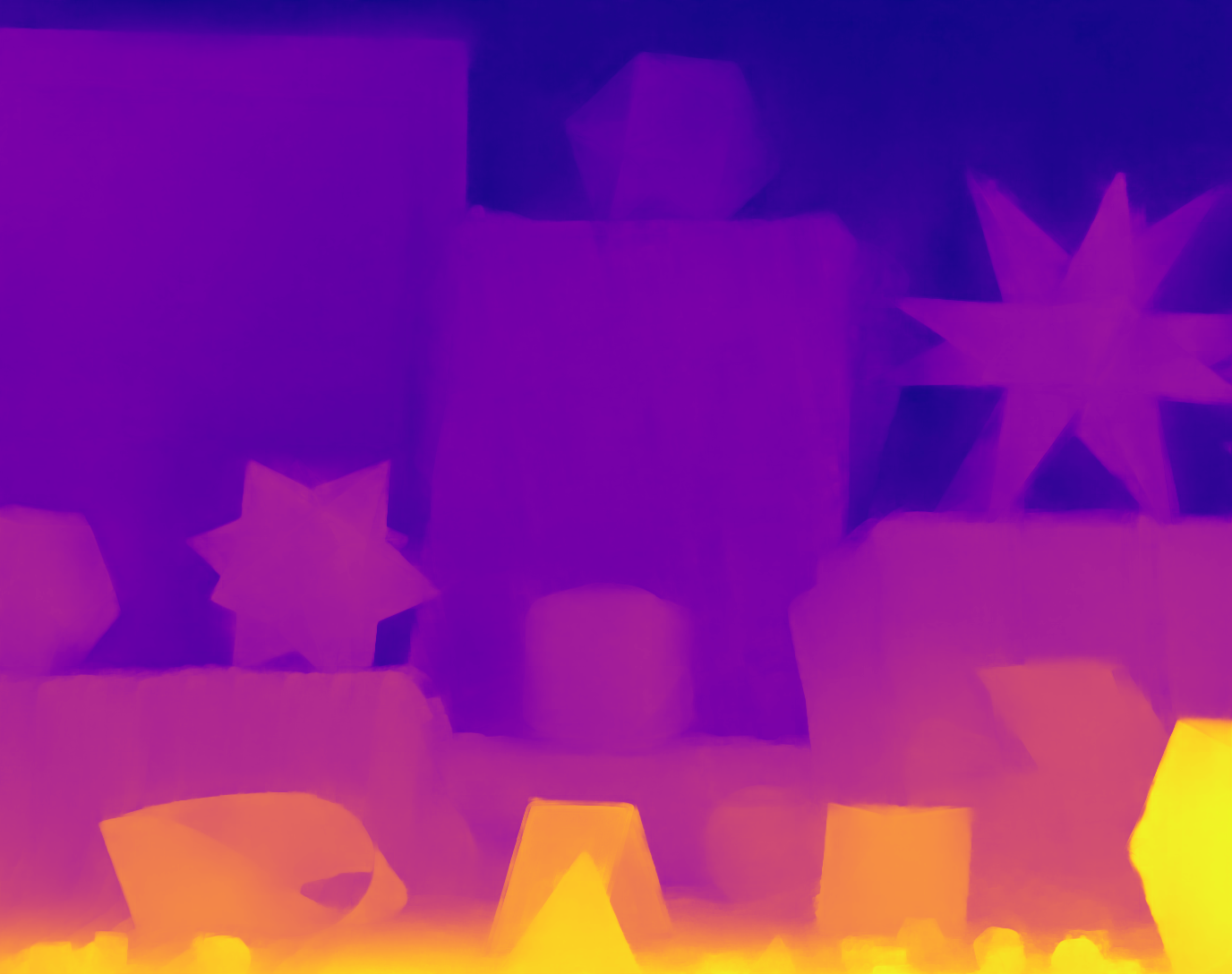}\vspace{-0.05cm}
    \end{minipage}    
  }
  \hspace{-0.15cm}
  \subfloat[{Ground truth}]{
    \begin{minipage}[b]{0.19\linewidth} 
      \includegraphics[width=\linewidth]{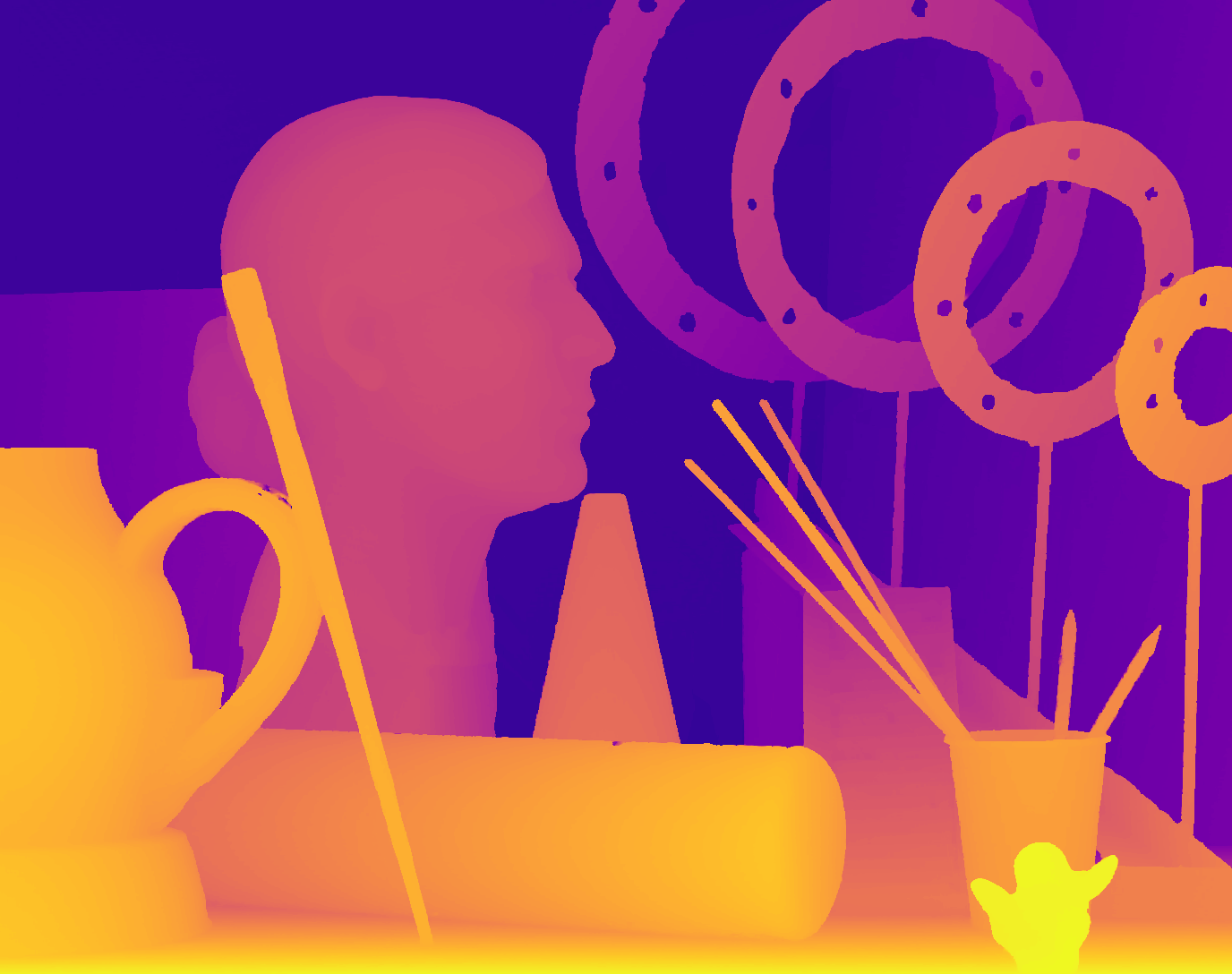}\vspace{-0.05cm}
      \includegraphics[width=\linewidth]{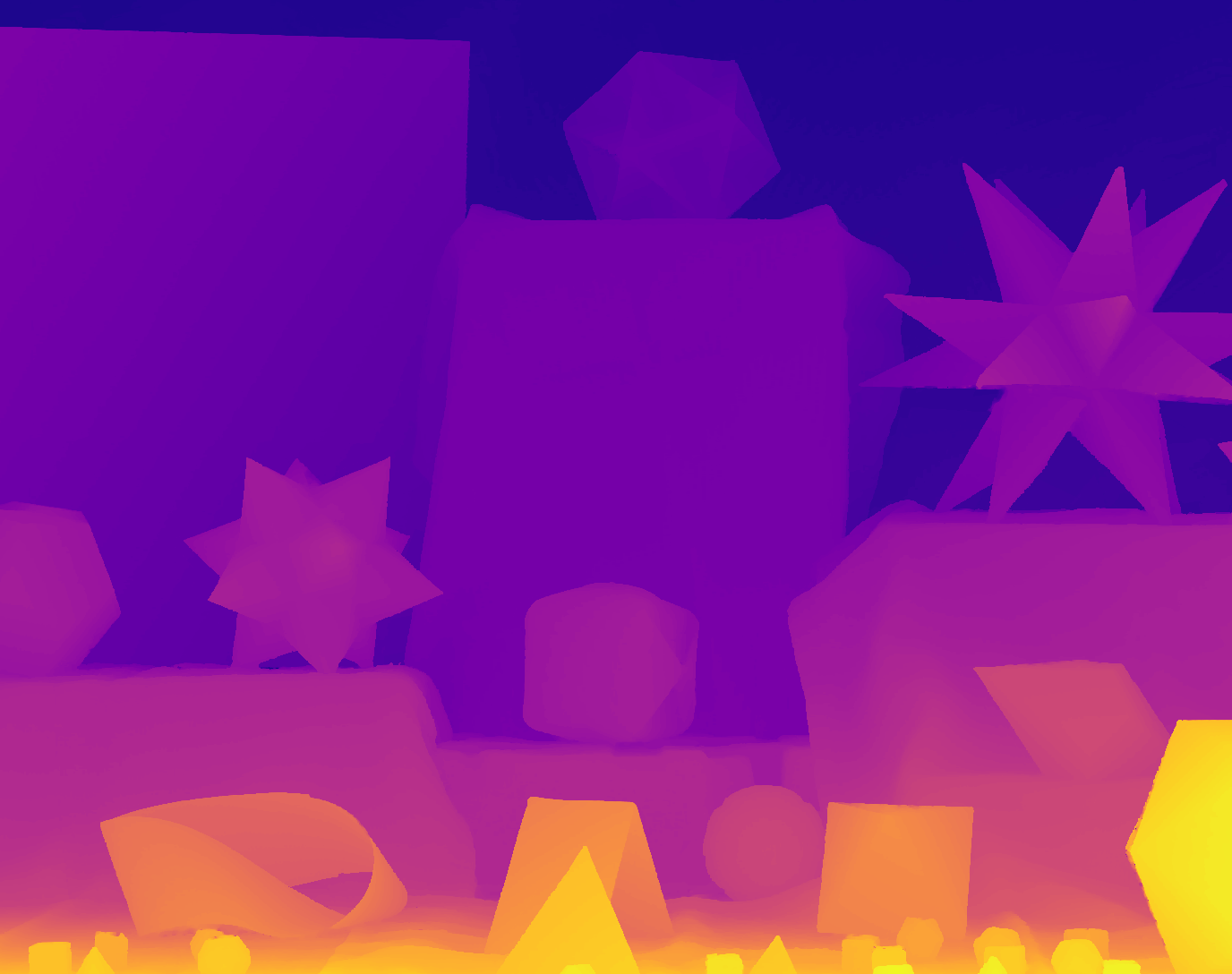}\vspace{-0.05cm}
    \end{minipage}    
  }
  \caption{Qualitative comparisons of guided noisy depth map super-resolution on the Noisy Middlebury dataset.
  The first row shows results of $\times8$ up-sampling on \textit{Art}, and the second row shows $\times16$ up-sampling on \textit{Moebius}.
  }
  \label{fig:noisy_depth_upsampling}
\end{figure*}

%% file: experiments.tex
In this section, we apply our method to guided depth map super-resolution and guided noisy depth map super-resolution tasks to demonstrate the effectiveness of our method.


\input{figs/mid_lu}

\subsection{Guided depth map super-resolution}

\subsubsection{Datasets and Evaluation Metrics} 
We adpot three widely-used benchmarks for the guided depth super-resolution task:
\begin{itemize}
\item NYU v2 dataset~\cite{Silberman2012IndoorSA}: This dataset provides 1449 RGBD pairs of indoor scenes captured by Microsoft Kinect~\cite{zhang2012kinect} using structural light. We use the first 1000 pairs as the training set and the rest 449 pairs as the evaluation set following previous work~\cite{Kim2021DeformableKN,Li2019JointIF}.
\item Middlebury dataset~\cite{Hirschmller2007EvaluationOC,Scharstein2007LearningCR}: we use a subset of 30 RGBD pairs from the 2001-2006 datasets provided by Lu et al.~\cite{lu2014depth} for testing. 
\item Lu dataset~\cite{lu2014depth}: This dataset consists of 6 RGBD pairs acquired by ASUS Xtion Pro camera. We use it for testing.
\end{itemize}
Following Kim et al.~\cite{Kim2021DeformableKN}, we train our model on the NYU v2 dataset, and test it on all the three datasets.
We do not fine-tune the model on Middlebury dataset or Lu dataset in order to test the generalization ability of the model.
The LR input images are generated at different ratios ($\times4$, $\times8$, $\times16$) through bicubic down-sampling from the HR target images.
We use average RMSE as the evaluation metric for the depth map super-resolution task.

\subsubsection{Implementation Details}
\label{sec:gdsr_impl}
We choose EDSR-baseline~\cite{Lim2017EnhancedDR} as the backbone for the two encoders, and discard the up-sampling modules to generate feature maps of the same size as the input image.
The output dimension of the encoder is set to $128$, and thus the input dimension of the DIF is $128 \times 3 + 2 = 386$.
A 5-layer MLP is used to model the DIF with decreasing hidden dimensions $(1024, 512, 256, 128)$.

We train the model for 200 epochs with the batch size of 1.
The HR image is randomly cropped into $(256, 256)$ patches and we sample $30720$ pixels per patch for each training step.
The depth maps are scaled to $[0, 1]$ before fed into the neural networks.
For the Middlebury dataset and the Lu dataset, we interpret the provided disparity map as the depth map according to~\cite{Kim2021DeformableKN}.
We use the Adam optimizer~\cite{kingma2015adam} to train our models.
The initial learning rate is set to $0.0001$ and is divided by $0.2$ for every $60$ epochs.
We apply data augmentation by flipping the image pairs vertically or horizontally in training.
When testing, all of the pixels in the HR domain are queried to recover the target image.
The model is implemented and trained using the \texttt{PyTorch} framework~\cite{paszke2017automatic}.

\subsubsection{Quantitative Comparisons}
We compare the proposed method with state-of-the-art methods, including recent learning based methods such as DJFR~\cite{Li2019JointIF} and DKN~\cite{Kim2021DeformableKN}. 
Table~\ref{tab:depth-upsampling} shows the detailed results on the three datasets.
We report the average RMSE on the test set. 
For the NYU v2 dataset, the average RMSE is measured in centimeters. 
For the Middlebury dataset and the Lu dataset, the average RMSE is measured in the original scale of the provided disparity.
Our method outperforms the existing methods by large margins in all datasets and settings.
With the proposed JIIF representation, our method predicts more accurate target image in all up-sampling ratios, and generalizes well into data from other sources (e.g., disparity maps acquired by different devices).
This improvement is from the strong capability of implicit neural representation and the joint leaning of interpolation values and weights.

\subsubsection{Qualitative Comparisons}
We provide visual comparison of the $\times8$ super-resolution results on the NYU v2 dataset in Figure~\ref{fig:depth_upsampling}. 
Also, generalization results on the Middlebury dataset and the Lu dataset are shown in Figure~\ref{fig:depth_upsampling_2}.
Our method produces more accurate and sharper edges in areas of complicated structures, where other methods fail to model the geometry and generate blurred results.
Besides, our method can restore reasonable structure even when the RGB guidance is ambiguous, e.g., too dark to provide any useful information.
This confirms the advantages of the proposed JIFF representation.

\input{tabs/abl_modules}
\input{tabs/abl_weights}

\subsection{Guided noisy depth map super-resolution}
To show the robustness of our method on noisy data, we further perform experiments to restore noisy low-resolution input depth maps to noise-free high-resolution target depth maps.

\subsubsection{Datasets} 
The Noisy Middlebury dataset~\cite{park2011} is used as the evaluation dataset for this task. 
It contains three standard RGBD pairs from the Middlebury 2005 dataset, i.e.,\textit{Art}, \textit{Books} and \textit{Moebius}.
We simulate noisy LR input following previous work~\cite{riegler16gdsr,Kim2021DeformableKN} by adding a conditional Gaussian noise to the LR input:
\begin{equation}
    n(x) \sim \mathcal N (0, \sigma x)
\end{equation}
where $x$ is proportional to the depth value (e.g., if the input is disparity $d$, we use $x = \frac 1 d$), and $\sigma$ is the magnitude of the noise.
For training, we use the NYU v2 dataset with the same type of noise added to the input images, and do not fine-tune the model on the Noisy Middlebury dataset.
In particular, the $sigma$ is set to $651$ for the Noisy Middlebury dataset following~\cite{riegler16gdsr}, and $0.04$ for the NYU v2 dataset to simulate similar magnitude of noise. 
The other experimental settings are the same as in Section~\ref{sec:gdsr_impl}.

\begin{figure}[ht]
    \centering
    \includegraphics[width=\linewidth]{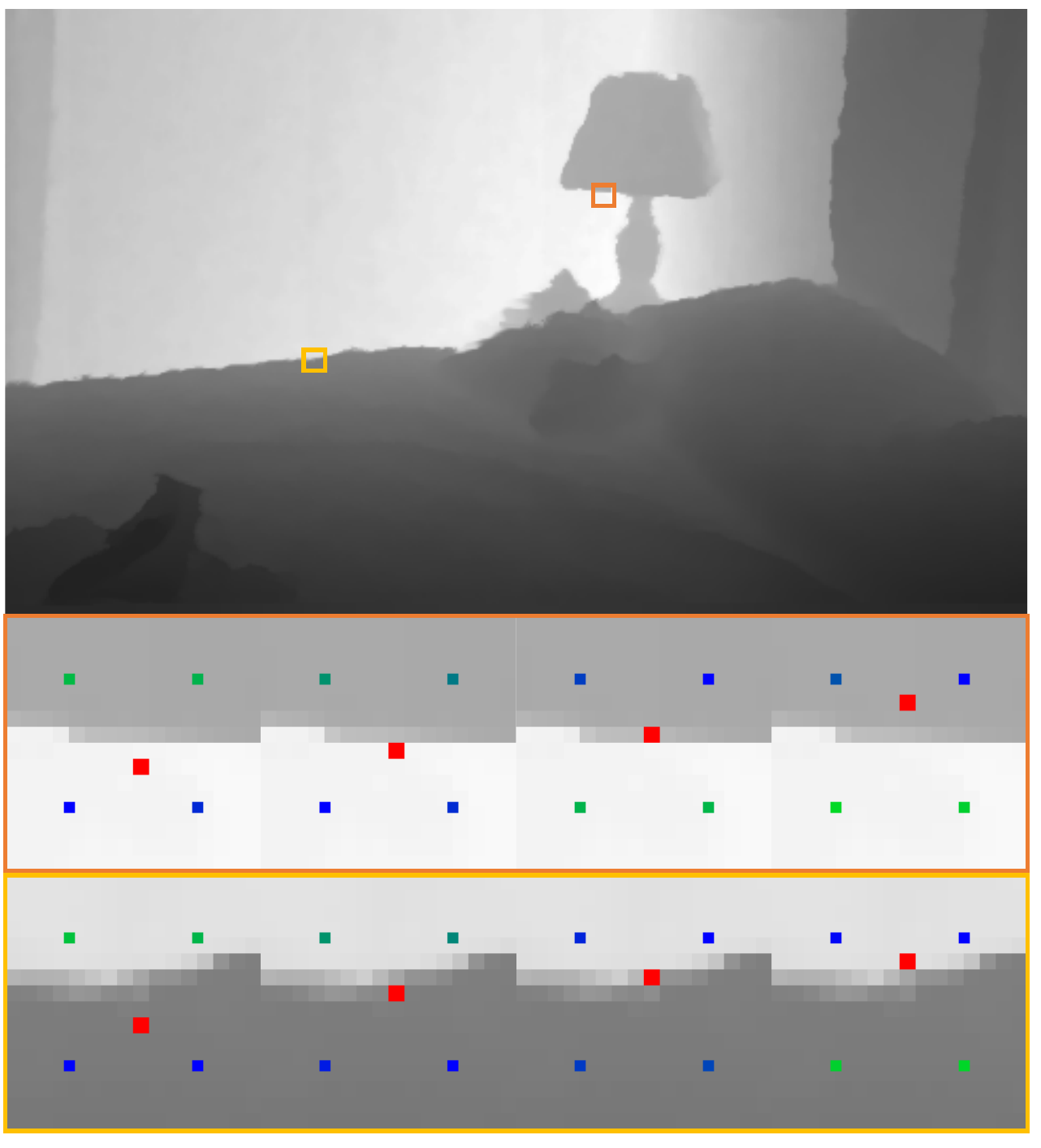}
    \caption{Visualization of the learned interpolation weights. 
    We show two examples of the query pixel crossing an edge in the HR target image.
    The query pixel is in red, and the four corner pixels' color indicates the learned interpolation weights.
    Higher weights are in bluer color, while lower weights are in greener color.}
    \label{fig:vis_edge}
\end{figure}

\subsubsection{Quantitative Comparisons} 
From Table~\ref{tab:noise-middlebury}, we can see our method outperforms other methods on most settings of the guided noisy depth map super-resolution task.
Although trained on depth maps from the NYU v2 dataset, our method generalizes well to the disparity maps from the Noisy Middlebury dataset.
This agrees with the previous experiments and demonstrates the noise suppression ability of our method.

\subsubsection{Qualitative Comparisons} 
The visual comparison of noisy depth map super-resolution is shown in Figure~\ref{fig:noisy_depth_upsampling}. 
Even the input image is corrupted severely by the noise and at $\times16$ up-sampling ratio, our method successfully restores reasonable structural details in the predicted HR depth map.
Also, our method can better suppress noises and restore sharper edges compared to other methods.


\subsection{Ablation Study}
We conduct ablation studies on different proposed modules in our method, and verify the effect of these modules for the $\times8$ guided depth super-resolution task on the NYU v2 dataset.

\subsubsection{Learning interpolation weights}
Firstly, we do ablation studies on different strategies to learn the interpolation weights.
From Table~\ref{tab:ablation_weights}, our graph attention based weights learning achieves the best performance.
`Bilinear' means the bilinear interpolation weights are used.
`Direct Regression' means we use directly a convolution layer to regress the weights from the guide image features, 
and `Graph Attention' means we apply a graph attention layer to regress the weights from the guide image features.
Compared to the baseline bilinear interpolation weights, our method reduces the average RMSE by $0.92$.
Direct regression of the weights used in DKN~\cite{Kim2021DeformableKN} also fails to learn meaningful interpolation weights.
This verifies the effectiveness of the graph attention mechanism for leaning edge weights.
We also provide the visualization of the learned interpolation weights in Figure~\ref{fig:vis_edge}.
The graph attention module can adapt to different locations dynamically, predicting higher weights if the two vertices share common guidance features.
For example, when the query pixel crosses an edge, the interpolation weights will switch to the correct side too.
This avoids assigning large weights to wrong values from the opposite side, which is one of the main causes of blur in traditional image interpolation.

\subsubsection{Joint learning of interpolation weights and values}
We argue that the interpolation weights and values are correlated and can be learned together to boost the performance.
Our JIFF representation is designed to exploit this correlation by predicting them in one DIF.
We conduct experiments to prove this hypothesis in Table~\ref{tab:ablation_modules}.
`Baseline' means that we break the JIIF into two DIFs that learn interpolation weights and values separately as described in Equation~\ref{eq:jiif_value} and~\ref{eq:jiif_weight}.
`Joint Repr.' means we use one unified MLP to learn interpolation weights and values as described in Equation~\ref{eq:jiif}.
Note that we use the same architecture for the DIFs, which means the `Joint Repr.' setting also reduces the last four MLP layers' parameters by half.
The experimental results validate our hypothesis that the joint representation can further enhance the final performance.

\subsubsection{Residual Learning}
Previous work~\cite{Li2019JointIF,Kim2021DeformableKN} has shown that residual learning, 
i.e. first up-sample the input with bicubic interpolation and then correct it by predicting the a residual image, 
can speed up convergence and improve the final performance.
We also adopt this idea and perform experiments to validate the effect of residual learning in Table~\ref{tab:ablation_modules}.
`Residual' means we adopt a residual learning framework.
With both residual learning and joint representation applied, our method achieves the best performance.

%% file: figs/mid_lu.tex
\begin{figure}[t]
  \centering
  \tiny	
  \subfloat[{}]{
    \begin{minipage}[b]{0.16\linewidth}
      \includegraphics[width=\linewidth]{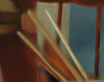}\vspace{-0.03cm}
      \includegraphics[width=\linewidth]{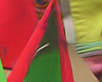}\vspace{-0.03cm}
      \includegraphics[width=\linewidth]{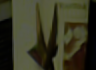}\vspace{-0.03cm}
    \end{minipage}
  }
  \hspace{-0.12cm}
  \subfloat[{}]{
    \begin{minipage}[b]{0.16\linewidth} 
      \includegraphics[width=\linewidth]{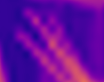}\vspace{-0.03cm}
      \includegraphics[width=\linewidth]{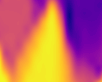}\vspace{-0.03cm}
      \includegraphics[width=\linewidth]{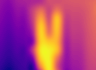}\vspace{-0.03cm}
    \end{minipage}    
  }
  \hspace{-0.12cm}
  \subfloat[{}]{
    \begin{minipage}[b]{0.16\linewidth} 
      \includegraphics[width=\linewidth]{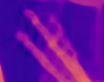}\vspace{-0.03cm}
      \includegraphics[width=\linewidth]{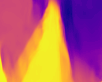}\vspace{-0.03cm}
      \includegraphics[width=\linewidth]{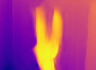}\vspace{-0.03cm}
    \end{minipage}    
  }
  \hspace{-0.12cm}
  \subfloat[{}]{
    \begin{minipage}[b]{0.16\linewidth} 
      \includegraphics[width=\linewidth]{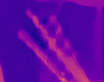}\vspace{-0.03cm}
      \includegraphics[width=\linewidth]{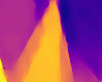}\vspace{-0.03cm}
      \includegraphics[width=\linewidth]{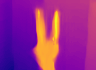}\vspace{-0.03cm}
    \end{minipage}    
  }
  \hspace{-0.12cm}
  \subfloat[{}]{
    \begin{minipage}[b]{0.16\linewidth} 
      \includegraphics[width=\linewidth]{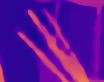}\vspace{-0.03cm}
      \includegraphics[width=\linewidth]{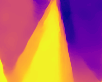}\vspace{-0.03cm}
      \includegraphics[width=\linewidth]{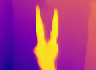}\vspace{-0.03cm}
    \end{minipage}    
  }
  \hspace{-0.12cm}
  \subfloat[{}]{
    \begin{minipage}[b]{0.16\linewidth} 
      \includegraphics[width=\linewidth]{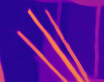}\vspace{-0.03cm}
      \includegraphics[width=\linewidth]{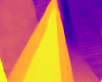}\vspace{-0.03cm}
      \includegraphics[width=\linewidth]{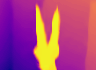}\vspace{-0.03cm}
    \end{minipage}    
  }
  \caption{Qualitative comparisons of $\times8$ guided depth map super-resolution on the Middlebury dataset (first two rows) and the Lu dataset (last row). From left to right: RGB images, Bicubic interpolation results, DJFR~\cite{Li2019JointIF} results, DKN~\cite{Kim2021DeformableKN} results, Our results, and the Ground Truth.}
  \label{fig:depth_upsampling_2}
  
\end{figure}

%% file: tabs/abl_modules.tex
\begin{table}[ht]
    \centering
    \caption{Ablation study on proposed modules.}
    \label{tab:ablation_modules}
    \newcolumntype{L}[1]{>{\raggedright\arraybackslash}p{#1}}
    \newcolumntype{C}[1]{>{\centering\arraybackslash}p{#1}}
    \newcolumntype{R}[1]{>{\raggedleft\arraybackslash}p{#1}}
    \setlength{\tabcolsep}{8pt}
    \begin{tabular}{C{1.4cm} C{1.4cm} C{1.4cm} C{1.4cm}}
        \shline
        Baseline   & Joint Repr. & Residual  & RMSE \\
        \hline
        \checkmark &            &            & 3.12 \\
        \checkmark & \checkmark &            & 2.95 \\
        \checkmark &            & \checkmark & 2.97 \\
        \checkmark & \checkmark & \checkmark & \textbf{2.76} \\
        \shline
    \end{tabular}
    \vspace{-0.3cm}
\end{table}

%% file: tabs/abl_weights.tex
\begin{table}[ht]
    \centering
    \caption{Ablation study on interpolation weights learning strategies.}
    \label{tab:ablation_weights}
    \newcolumntype{L}[1]{>{\raggedright\arraybackslash}p{#1}}
    \newcolumntype{C}[1]{>{\centering\arraybackslash}p{#1}}
    \newcolumntype{R}[1]{>{\raggedleft\arraybackslash}p{#1}}
    \begin{tabular}{L{4.5cm} C{2cm}}
        \shline
        Methods & RMSE \\
        \hline
        Bilinear & 3.68 \\
        Direct Regression & 3.67 \\
        Graph Attention & \textbf{2.76} \\
        \shline
    \end{tabular}
    \vspace{-0.3cm}
\end{table}

%% file: conclusion.tex
In this paper, we propose a Joint Implicit Image Function (JIIF) representation for the guided super-resolution task.
In JIIF representation, we take the form of a general image interpolation problem but equip it with deep implicit functions to enhance the model capability.
In particular, the target image is represented with spatially distributed local latent codes extracted from both the input image and the guide image, and we use a graph attention mechanism to learn the interpolation weights at the same time in one unified deep implicit function.
We demonstrate that the learning of interpolation weights and values are correlated, and our JIIF representation takes advantage of this correlation to boost performance.
The effectiveness and generalization ability of our method is verified on two tasks.

%% file: main.bbl

\begin{thebibliography}{59}


\ifx \showCODEN    \undefined \def \showCODEN     #1{\unskip}     \fi
\ifx \showDOI      \undefined \def \showDOI       #1{#1}\fi
\ifx \showISBNx    \undefined \def \showISBNx     #1{\unskip}     \fi
\ifx \showISBNxiii \undefined \def \showISBNxiii  #1{\unskip}     \fi
\ifx \showISSN     \undefined \def \showISSN      #1{\unskip}     \fi
\ifx \showLCCN     \undefined \def \showLCCN      #1{\unskip}     \fi
\ifx \shownote     \undefined \def \shownote      #1{#1}          \fi
\ifx \showarticletitle \undefined \def \showarticletitle #1{#1}   \fi
\ifx \showURL      \undefined \def \showURL       {\relax}        \fi
\providecommand\bibfield[2]{#2}
\providecommand\bibinfo[2]{#2}
\providecommand\natexlab[1]{#1}
\providecommand\showeprint[2][]{arXiv:#2}

\bibitem[\protect\citeauthoryear{Barron and Poole}{Barron and Poole}{2016}]%
        {barron2016fast}
\bibfield{author}{\bibinfo{person}{Jonathan~T Barron} {and}
  \bibinfo{person}{Ben Poole}.} \bibinfo{year}{2016}\natexlab{}.
\newblock \showarticletitle{The fast bilateral solver}. In
  \bibinfo{booktitle}{\emph{European Conference on Computer Vision}}. Springer,
  \bibinfo{pages}{617--632}.
\newblock


\bibitem[\protect\citeauthoryear{Chan, Buisman, Theobalt, and Thrun}{Chan
  et~al\mbox{.}}{2008}]%
        {Chan2008ANF}
\bibfield{author}{\bibinfo{person}{Derek Chan}, \bibinfo{person}{Hylke
  Buisman}, \bibinfo{person}{Christian Theobalt}, {and}
  \bibinfo{person}{Sebastian Thrun}.} \bibinfo{year}{2008}\natexlab{}.
\newblock \showarticletitle{A noise-aware filter for real-time depth
  upsampling}. In \bibinfo{booktitle}{\emph{Workshop on Multi-camera and
  Multi-modal Sensor Fusion Algorithms and Applications-M2SFA2 2008}}.
\newblock


\bibitem[\protect\citeauthoryear{Chang, Yeung, and Xiong}{Chang
  et~al\mbox{.}}{2004}]%
        {chang2004super}
\bibfield{author}{\bibinfo{person}{Hong Chang}, \bibinfo{person}{Dit-Yan
  Yeung}, {and} \bibinfo{person}{Yimin Xiong}.}
  \bibinfo{year}{2004}\natexlab{}.
\newblock \showarticletitle{Super-resolution through neighbor embedding}. In
  \bibinfo{booktitle}{\emph{Proceedings of the 2004 IEEE Computer Society
  Conference on Computer Vision and Pattern Recognition, 2004. CVPR 2004.}},
  Vol.~\bibinfo{volume}{1}. IEEE, \bibinfo{pages}{I--I}.
\newblock


\bibitem[\protect\citeauthoryear{Chen, Lin, Qian, Zeng, and Li}{Chen
  et~al\mbox{.}}{2020a}]%
        {chen20203d}
\bibfield{author}{\bibinfo{person}{Xiaokang Chen}, \bibinfo{person}{Kwan-Yee
  Lin}, \bibinfo{person}{Chen Qian}, \bibinfo{person}{Gang Zeng}, {and}
  \bibinfo{person}{Hongsheng Li}.} \bibinfo{year}{2020}\natexlab{a}.
\newblock \showarticletitle{3d sketch-aware semantic scene completion via
  semi-supervised structure prior}. In \bibinfo{booktitle}{\emph{Proceedings of
  the IEEE/CVF Conference on Computer Vision and Pattern Recognition}}.
  \bibinfo{pages}{4193--4202}.
\newblock


\bibitem[\protect\citeauthoryear{Chen, Lin, Wang, Wu, Qian, Li, and Zeng}{Chen
  et~al\mbox{.}}{2020b}]%
        {chen2020bi}
\bibfield{author}{\bibinfo{person}{Xiaokang Chen}, \bibinfo{person}{Kwan-Yee
  Lin}, \bibinfo{person}{Jingbo Wang}, \bibinfo{person}{Wayne Wu},
  \bibinfo{person}{Chen Qian}, \bibinfo{person}{Hongsheng Li}, {and}
  \bibinfo{person}{Gang Zeng}.} \bibinfo{year}{2020}\natexlab{b}.
\newblock \showarticletitle{Bi-directional cross-modality feature propagation
  with separation-and-aggregation gate for RGB-D semantic segmentation}. In
  \bibinfo{booktitle}{\emph{Computer Vision--ECCV 2020: 16th European
  Conference, Glasgow, UK, August 23--28, 2020, Proceedings, Part XI 16}}.
  Springer, \bibinfo{pages}{561--577}.
\newblock


\bibitem[\protect\citeauthoryear{Chen, Xing, and Zeng}{Chen
  et~al\mbox{.}}{2020c}]%
        {chen2020real}
\bibfield{author}{\bibinfo{person}{Xiaokang Chen}, \bibinfo{person}{Yajie
  Xing}, {and} \bibinfo{person}{Gang Zeng}.} \bibinfo{year}{2020}\natexlab{c}.
\newblock \showarticletitle{Real-Time Semantic Scene Completion Via Feature
  Aggregation And Conditioned Prediction}. In \bibinfo{booktitle}{\emph{2020
  IEEE International Conference on Image Processing (ICIP)}}. IEEE,
  \bibinfo{pages}{2830--2834}.
\newblock


\bibitem[\protect\citeauthoryear{Chen, Liu, and Wang}{Chen
  et~al\mbox{.}}{2021}]%
        {Chen2020LearningCI}
\bibfield{author}{\bibinfo{person}{Yinbo Chen}, \bibinfo{person}{Sifei Liu},
  {and} \bibinfo{person}{Xiaolong Wang}.} \bibinfo{year}{2021}\natexlab{}.
\newblock \showarticletitle{Learning continuous image representation with local
  implicit image function}. In \bibinfo{booktitle}{\emph{Proceedings of the
  IEEE/CVF Conference on Computer Vision and Pattern Recognition}}.
  \bibinfo{pages}{8628--8638}.
\newblock


\bibitem[\protect\citeauthoryear{Diebel and Thrun}{Diebel and Thrun}{2005}]%
        {diebel2006application}
\bibfield{author}{\bibinfo{person}{James Diebel} {and}
  \bibinfo{person}{Sebastian Thrun}.} \bibinfo{year}{2005}\natexlab{}.
\newblock \showarticletitle{An application of markov random fields to range
  sensing}. In \bibinfo{booktitle}{\emph{NIPS}}, Vol.~\bibinfo{volume}{5}.
  \bibinfo{pages}{291--298}.
\newblock


\bibitem[\protect\citeauthoryear{Dong, Loy, He, and Tang}{Dong
  et~al\mbox{.}}{2014}]%
        {Dong2014LearningAD}
\bibfield{author}{\bibinfo{person}{Chao Dong}, \bibinfo{person}{Chen~Change
  Loy}, \bibinfo{person}{Kaiming He}, {and} \bibinfo{person}{Xiaoou Tang}.}
  \bibinfo{year}{2014}\natexlab{}.
\newblock \showarticletitle{Learning a deep convolutional network for image
  super-resolution}. In \bibinfo{booktitle}{\emph{European conference on
  computer vision}}. Springer, \bibinfo{pages}{184--199}.
\newblock


\bibitem[\protect\citeauthoryear{Ferstl, Reinbacher, Ranftl, R{\"u}ther, and
  Bischof}{Ferstl et~al\mbox{.}}{2013}]%
        {ferstl2013}
\bibfield{author}{\bibinfo{person}{David Ferstl}, \bibinfo{person}{Christian
  Reinbacher}, \bibinfo{person}{Rene Ranftl}, \bibinfo{person}{Matthias
  R{\"u}ther}, {and} \bibinfo{person}{Horst Bischof}.}
  \bibinfo{year}{2013}\natexlab{}.
\newblock \showarticletitle{Image guided depth upsampling using anisotropic
  total generalized variation}. In \bibinfo{booktitle}{\emph{Proceedings of the
  IEEE International Conference on Computer Vision}}.
  \bibinfo{pages}{993--1000}.
\newblock


\bibitem[\protect\citeauthoryear{Genova, Cole, Sud, Sarna, and
  Funkhouser}{Genova et~al\mbox{.}}{2020}]%
        {Genova2020LocalDI}
\bibfield{author}{\bibinfo{person}{Kyle Genova}, \bibinfo{person}{Forrester
  Cole}, \bibinfo{person}{Avneesh Sud}, \bibinfo{person}{Aaron Sarna}, {and}
  \bibinfo{person}{Thomas Funkhouser}.} \bibinfo{year}{2020}\natexlab{}.
\newblock \showarticletitle{Local deep implicit functions for 3d shape}. In
  \bibinfo{booktitle}{\emph{Proceedings of the IEEE/CVF Conference on Computer
  Vision and Pattern Recognition}}. \bibinfo{pages}{4857--4866}.
\newblock


\bibitem[\protect\citeauthoryear{Gu, Zuo, Guo, Chen, Chen, and Zhang}{Gu
  et~al\mbox{.}}{2017}]%
        {Gu2017learning}
\bibfield{author}{\bibinfo{person}{Shuhang Gu}, \bibinfo{person}{Wangmeng Zuo},
  \bibinfo{person}{Shi Guo}, \bibinfo{person}{Yunjin Chen},
  \bibinfo{person}{Chongyu Chen}, {and} \bibinfo{person}{Lei Zhang}.}
  \bibinfo{year}{2017}\natexlab{}.
\newblock \showarticletitle{Learning dynamic guidance for depth image
  enhancement}. In \bibinfo{booktitle}{\emph{Proceedings of the IEEE conference
  on computer vision and pattern recognition}}. \bibinfo{pages}{3769--3778}.
\newblock


\bibitem[\protect\citeauthoryear{Gupta, Girshick, Arbel{\'a}ez, and
  Malik}{Gupta et~al\mbox{.}}{2014}]%
        {gupta2014learning}
\bibfield{author}{\bibinfo{person}{Saurabh Gupta}, \bibinfo{person}{Ross
  Girshick}, \bibinfo{person}{Pablo Arbel{\'a}ez}, {and}
  \bibinfo{person}{Jitendra Malik}.} \bibinfo{year}{2014}\natexlab{}.
\newblock \showarticletitle{Learning rich features from RGB-D images for object
  detection and segmentation}. In \bibinfo{booktitle}{\emph{European conference
  on computer vision}}. Springer, \bibinfo{pages}{345--360}.
\newblock


\bibitem[\protect\citeauthoryear{Ham, Cho, and Ponce}{Ham
  et~al\mbox{.}}{2017}]%
        {ham2018robust}
\bibfield{author}{\bibinfo{person}{Bumsub Ham}, \bibinfo{person}{Minsu Cho},
  {and} \bibinfo{person}{Jean Ponce}.} \bibinfo{year}{2017}\natexlab{}.
\newblock \showarticletitle{Robust guided image filtering using nonconvex
  potentials}.
\newblock \bibinfo{journal}{\emph{IEEE transactions on pattern analysis and
  machine intelligence}} \bibinfo{volume}{40}, \bibinfo{number}{1}
  (\bibinfo{year}{2017}), \bibinfo{pages}{192--207}.
\newblock


\bibitem[\protect\citeauthoryear{He, Sun, and Tang}{He et~al\mbox{.}}{2012}]%
        {he2013guided}
\bibfield{author}{\bibinfo{person}{Kaiming He}, \bibinfo{person}{Jian Sun},
  {and} \bibinfo{person}{Xiaoou Tang}.} \bibinfo{year}{2012}\natexlab{}.
\newblock \showarticletitle{Guided image filtering}.
\newblock \bibinfo{journal}{\emph{IEEE transactions on pattern analysis and
  machine intelligence}} \bibinfo{volume}{35}, \bibinfo{number}{6}
  (\bibinfo{year}{2012}), \bibinfo{pages}{1397--1409}.
\newblock


\bibitem[\protect\citeauthoryear{Hirschmuller and Scharstein}{Hirschmuller and
  Scharstein}{2007}]%
        {Hirschmller2007EvaluationOC}
\bibfield{author}{\bibinfo{person}{Heiko Hirschmuller} {and}
  \bibinfo{person}{Daniel Scharstein}.} \bibinfo{year}{2007}\natexlab{}.
\newblock \showarticletitle{Evaluation of cost functions for stereo matching}.
  In \bibinfo{booktitle}{\emph{2007 IEEE Conference on Computer Vision and
  Pattern Recognition}}. IEEE, \bibinfo{pages}{1--8}.
\newblock


\bibitem[\protect\citeauthoryear{Hui, Loy, and Tang}{Hui et~al\mbox{.}}{2016}]%
        {Hui2016DepthMS}
\bibfield{author}{\bibinfo{person}{Tak-Wai Hui}, \bibinfo{person}{Chen~Change
  Loy}, {and} \bibinfo{person}{Xiaoou Tang}.} \bibinfo{year}{2016}\natexlab{}.
\newblock \showarticletitle{Depth map super-resolution by deep multi-scale
  guidance}. In \bibinfo{booktitle}{\emph{European conference on computer
  vision}}. Springer, \bibinfo{pages}{353--369}.
\newblock


\bibitem[\protect\citeauthoryear{Jiang, Sud, Makadia, Huang, Nie{\ss}ner,
  Funkhouser, et~al\mbox{.}}{Jiang et~al\mbox{.}}{2020}]%
        {Jiang2020LocalIG}
\bibfield{author}{\bibinfo{person}{Chiyu Jiang}, \bibinfo{person}{Avneesh Sud},
  \bibinfo{person}{Ameesh Makadia}, \bibinfo{person}{Jingwei Huang},
  \bibinfo{person}{Matthias Nie{\ss}ner}, \bibinfo{person}{Thomas Funkhouser},
  {et~al\mbox{.}}} \bibinfo{year}{2020}\natexlab{}.
\newblock \showarticletitle{Local implicit grid representations for 3d scenes}.
  In \bibinfo{booktitle}{\emph{Proceedings of the IEEE/CVF Conference on
  Computer Vision and Pattern Recognition}}. \bibinfo{pages}{6001--6010}.
\newblock


\bibitem[\protect\citeauthoryear{Kiechle, Hawe, and Kleinsteuber}{Kiechle
  et~al\mbox{.}}{2013}]%
        {Kiechle2013AJI}
\bibfield{author}{\bibinfo{person}{Martin Kiechle}, \bibinfo{person}{S. Hawe},
  {and} \bibinfo{person}{M. Kleinsteuber}.} \bibinfo{year}{2013}\natexlab{}.
\newblock \showarticletitle{A Joint Intensity and Depth Co-sparse Analysis
  Model for Depth Map Super-resolution}. In \bibinfo{booktitle}{\emph{2013 IEEE
  International Conference on Computer Vision}}. \bibinfo{pages}{1545--1552}.
\newblock


\bibitem[\protect\citeauthoryear{Kim, Ponce, and Ham}{Kim
  et~al\mbox{.}}{2021}]%
        {Kim2021DeformableKN}
\bibfield{author}{\bibinfo{person}{Beomjun Kim}, \bibinfo{person}{Jean Ponce},
  {and} \bibinfo{person}{Bumsub Ham}.} \bibinfo{year}{2021}\natexlab{}.
\newblock \showarticletitle{Deformable kernel networks for joint image
  filtering}.
\newblock \bibinfo{journal}{\emph{International Journal of Computer Vision}}
  \bibinfo{volume}{129}, \bibinfo{number}{2} (\bibinfo{year}{2021}),
  \bibinfo{pages}{579--600}.
\newblock


\bibitem[\protect\citeauthoryear{Kingma and Ba}{Kingma and Ba}{2014}]%
        {kingma2015adam}
\bibfield{author}{\bibinfo{person}{Diederik~P Kingma} {and}
  \bibinfo{person}{Jimmy Ba}.} \bibinfo{year}{2014}\natexlab{}.
\newblock \showarticletitle{Adam: A method for stochastic optimization}.
\newblock \bibinfo{journal}{\emph{arXiv preprint arXiv:1412.6980}}
  (\bibinfo{year}{2014}).
\newblock


\bibitem[\protect\citeauthoryear{Kipf and Welling}{Kipf and Welling}{2016}]%
        {Kipf2017SemiSupervisedCW}
\bibfield{author}{\bibinfo{person}{Thomas~N Kipf} {and} \bibinfo{person}{Max
  Welling}.} \bibinfo{year}{2016}\natexlab{}.
\newblock \showarticletitle{Semi-supervised classification with graph
  convolutional networks}.
\newblock \bibinfo{journal}{\emph{arXiv preprint arXiv:1609.02907}}
  (\bibinfo{year}{2016}).
\newblock


\bibitem[\protect\citeauthoryear{Kopf, Cohen, Lischinski, and Uyttendaele}{Kopf
  et~al\mbox{.}}{2007}]%
        {kopf2007joint}
\bibfield{author}{\bibinfo{person}{Johannes Kopf}, \bibinfo{person}{Michael~F
  Cohen}, \bibinfo{person}{Dani Lischinski}, {and} \bibinfo{person}{Matt
  Uyttendaele}.} \bibinfo{year}{2007}\natexlab{}.
\newblock \showarticletitle{Joint bilateral upsampling}.
\newblock \bibinfo{journal}{\emph{ACM Transactions on Graphics (ToG)}}
  \bibinfo{volume}{26}, \bibinfo{number}{3} (\bibinfo{year}{2007}),
  \bibinfo{pages}{96--es}.
\newblock


\bibitem[\protect\citeauthoryear{Lai, Huang, Ahuja, and Yang}{Lai
  et~al\mbox{.}}{2017}]%
        {Lai2017DeepLP}
\bibfield{author}{\bibinfo{person}{Wei-Sheng Lai}, \bibinfo{person}{Jia-Bin
  Huang}, \bibinfo{person}{Narendra Ahuja}, {and} \bibinfo{person}{Ming-Hsuan
  Yang}.} \bibinfo{year}{2017}\natexlab{}.
\newblock \showarticletitle{Deep laplacian pyramid networks for fast and
  accurate super-resolution}. In \bibinfo{booktitle}{\emph{Proceedings of the
  IEEE conference on computer vision and pattern recognition}}.
  \bibinfo{pages}{624--632}.
\newblock


\bibitem[\protect\citeauthoryear{Ledig, Theis, Husz{\'a}r, Caballero,
  Cunningham, Acosta, Aitken, Tejani, Totz, Wang, et~al\mbox{.}}{Ledig
  et~al\mbox{.}}{2017}]%
        {Ledig2017PhotoRealisticSI}
\bibfield{author}{\bibinfo{person}{Christian Ledig}, \bibinfo{person}{Lucas
  Theis}, \bibinfo{person}{Ferenc Husz{\'a}r}, \bibinfo{person}{Jose
  Caballero}, \bibinfo{person}{Andrew Cunningham}, \bibinfo{person}{Alejandro
  Acosta}, \bibinfo{person}{Andrew Aitken}, \bibinfo{person}{Alykhan Tejani},
  \bibinfo{person}{Johannes Totz}, \bibinfo{person}{Zehan Wang},
  {et~al\mbox{.}}} \bibinfo{year}{2017}\natexlab{}.
\newblock \showarticletitle{Photo-realistic single image super-resolution using
  a generative adversarial network}. In \bibinfo{booktitle}{\emph{Proceedings
  of the IEEE conference on computer vision and pattern recognition}}.
  \bibinfo{pages}{4681--4690}.
\newblock


\bibitem[\protect\citeauthoryear{Li, Huang, Ahuja, and Yang}{Li
  et~al\mbox{.}}{2016}]%
        {li2016deep}
\bibfield{author}{\bibinfo{person}{Yijun Li}, \bibinfo{person}{Jia-Bin Huang},
  \bibinfo{person}{Narendra Ahuja}, {and} \bibinfo{person}{Ming-Hsuan Yang}.}
  \bibinfo{year}{2016}\natexlab{}.
\newblock \showarticletitle{Deep joint image filtering}. In
  \bibinfo{booktitle}{\emph{European Conference on Computer Vision}}. Springer,
  \bibinfo{pages}{154--169}.
\newblock


\bibitem[\protect\citeauthoryear{Li, Huang, Ahuja, and Yang}{Li
  et~al\mbox{.}}{2019}]%
        {Li2019JointIF}
\bibfield{author}{\bibinfo{person}{Yijun Li}, \bibinfo{person}{Jia-Bin Huang},
  \bibinfo{person}{Narendra Ahuja}, {and} \bibinfo{person}{Ming-Hsuan Yang}.}
  \bibinfo{year}{2019}\natexlab{}.
\newblock \showarticletitle{Joint image filtering with deep convolutional
  networks}.
\newblock \bibinfo{journal}{\emph{IEEE transactions on pattern analysis and
  machine intelligence}} \bibinfo{volume}{41}, \bibinfo{number}{8}
  (\bibinfo{year}{2019}), \bibinfo{pages}{1909--1923}.
\newblock


\bibitem[\protect\citeauthoryear{Li, Xue, Sun, and Liu}{Li
  et~al\mbox{.}}{2012}]%
        {Li2012JointED}
\bibfield{author}{\bibinfo{person}{Yanjie Li}, \bibinfo{person}{Tianfan Xue},
  \bibinfo{person}{Lifeng Sun}, {and} \bibinfo{person}{Jianzhuang Liu}.}
  \bibinfo{year}{2012}\natexlab{}.
\newblock \showarticletitle{Joint Example-Based Depth Map Super-Resolution}. In
  \bibinfo{booktitle}{\emph{2012 IEEE International Conference on Multimedia
  and Expo}}. \bibinfo{pages}{152--157}.
\newblock


\bibitem[\protect\citeauthoryear{Lim, Son, Kim, Nah, and Mu~Lee}{Lim
  et~al\mbox{.}}{2017}]%
        {Lim2017EnhancedDR}
\bibfield{author}{\bibinfo{person}{Bee Lim}, \bibinfo{person}{Sanghyun Son},
  \bibinfo{person}{Heewon Kim}, \bibinfo{person}{Seungjun Nah}, {and}
  \bibinfo{person}{Kyoung Mu~Lee}.} \bibinfo{year}{2017}\natexlab{}.
\newblock \showarticletitle{Enhanced deep residual networks for single image
  super-resolution}. In \bibinfo{booktitle}{\emph{Proceedings of the IEEE
  conference on computer vision and pattern recognition workshops}}.
  \bibinfo{pages}{136--144}.
\newblock


\bibitem[\protect\citeauthoryear{Lu, Ren, and Liu}{Lu et~al\mbox{.}}{2014}]%
        {lu2014depth}
\bibfield{author}{\bibinfo{person}{Si Lu}, \bibinfo{person}{Xiaofeng Ren},
  {and} \bibinfo{person}{Feng Liu}.} \bibinfo{year}{2014}\natexlab{}.
\newblock \showarticletitle{Depth enhancement via low-rank matrix completion}.
  In \bibinfo{booktitle}{\emph{Proceedings of the IEEE conference on computer
  vision and pattern recognition}}. \bibinfo{pages}{3390--3397}.
\newblock


\bibitem[\protect\citeauthoryear{Lutio, D'aronco, Wegner, and Schindler}{Lutio
  et~al\mbox{.}}{2019}]%
        {Lutio2019GuidedSA}
\bibfield{author}{\bibinfo{person}{Riccardo~de Lutio}, \bibinfo{person}{Stefano
  D'aronco}, \bibinfo{person}{Jan~Dirk Wegner}, {and} \bibinfo{person}{Konrad
  Schindler}.} \bibinfo{year}{2019}\natexlab{}.
\newblock \showarticletitle{Guided super-resolution as pixel-to-pixel
  transformation}. In \bibinfo{booktitle}{\emph{Proceedings of the IEEE/CVF
  International Conference on Computer Vision}}. \bibinfo{pages}{8829--8837}.
\newblock


\bibitem[\protect\citeauthoryear{Ma, He, Wei, Sun, and Wu}{Ma
  et~al\mbox{.}}{2013}]%
        {Ma2013ConstantTW}
\bibfield{author}{\bibinfo{person}{Ziyang Ma}, \bibinfo{person}{Kaiming He},
  \bibinfo{person}{Yichen Wei}, \bibinfo{person}{Jian Sun}, {and}
  \bibinfo{person}{Enhua Wu}.} \bibinfo{year}{2013}\natexlab{}.
\newblock \showarticletitle{Constant time weighted median filtering for stereo
  matching and beyond}. In \bibinfo{booktitle}{\emph{Proceedings of the IEEE
  International Conference on Computer Vision}}. \bibinfo{pages}{49--56}.
\newblock


\bibitem[\protect\citeauthoryear{Mescheder, Oechsle, Niemeyer, Nowozin, and
  Geiger}{Mescheder et~al\mbox{.}}{2019}]%
        {Mescheder2019OccupancyNL}
\bibfield{author}{\bibinfo{person}{Lars Mescheder}, \bibinfo{person}{Michael
  Oechsle}, \bibinfo{person}{Michael Niemeyer}, \bibinfo{person}{Sebastian
  Nowozin}, {and} \bibinfo{person}{Andreas Geiger}.}
  \bibinfo{year}{2019}\natexlab{}.
\newblock \showarticletitle{Occupancy networks: Learning 3d reconstruction in
  function space}. In \bibinfo{booktitle}{\emph{Proceedings of the IEEE/CVF
  Conference on Computer Vision and Pattern Recognition}}.
  \bibinfo{pages}{4460--4470}.
\newblock


\bibitem[\protect\citeauthoryear{Park, Kim, Tai, Brown, and Kweon}{Park
  et~al\mbox{.}}{2011}]%
        {park2011}
\bibfield{author}{\bibinfo{person}{Jaesik Park}, \bibinfo{person}{Hyeongwoo
  Kim}, \bibinfo{person}{Yu-Wing Tai}, \bibinfo{person}{Michael~S Brown}, {and}
  \bibinfo{person}{Inso Kweon}.} \bibinfo{year}{2011}\natexlab{}.
\newblock \showarticletitle{High quality depth map upsampling for 3d-tof
  cameras}. In \bibinfo{booktitle}{\emph{2011 International Conference on
  Computer Vision}}. IEEE, \bibinfo{pages}{1623--1630}.
\newblock


\bibitem[\protect\citeauthoryear{Park, Florence, Straub, Newcombe, and
  Lovegrove}{Park et~al\mbox{.}}{2019}]%
        {Park2019DeepSDFLC}
\bibfield{author}{\bibinfo{person}{Jeong~Joon Park}, \bibinfo{person}{Peter
  Florence}, \bibinfo{person}{Julian Straub}, \bibinfo{person}{Richard
  Newcombe}, {and} \bibinfo{person}{Steven Lovegrove}.}
  \bibinfo{year}{2019}\natexlab{}.
\newblock \showarticletitle{Deepsdf: Learning continuous signed distance
  functions for shape representation}. In \bibinfo{booktitle}{\emph{Proceedings
  of the IEEE/CVF Conference on Computer Vision and Pattern Recognition}}.
  \bibinfo{pages}{165--174}.
\newblock


\bibitem[\protect\citeauthoryear{Paszke, Gross, Chintala, Chanan, Yang, DeVito,
  Lin, Desmaison, Antiga, and Lerer}{Paszke et~al\mbox{.}}{2017}]%
        {paszke2017automatic}
\bibfield{author}{\bibinfo{person}{Adam Paszke}, \bibinfo{person}{Sam Gross},
  \bibinfo{person}{Soumith Chintala}, \bibinfo{person}{Gregory Chanan},
  \bibinfo{person}{Edward Yang}, \bibinfo{person}{Zachary DeVito},
  \bibinfo{person}{Zeming Lin}, \bibinfo{person}{Alban Desmaison},
  \bibinfo{person}{Luca Antiga}, {and} \bibinfo{person}{Adam Lerer}.}
  \bibinfo{year}{2017}\natexlab{}.
\newblock \showarticletitle{Automatic differentiation in pytorch}.
\newblock  (\bibinfo{year}{2017}).
\newblock


\bibitem[\protect\citeauthoryear{Riegler, Ferstl, R{\"u}ther, and
  Bischof}{Riegler et~al\mbox{.}}{2016}]%
        {riegler16gdsr}
\bibfield{author}{\bibinfo{person}{Gernot Riegler}, \bibinfo{person}{David
  Ferstl}, \bibinfo{person}{Matthias R{\"u}ther}, {and} \bibinfo{person}{Horst
  Bischof}.} \bibinfo{year}{2016}\natexlab{}.
\newblock \showarticletitle{A deep primal-dual network for guided depth
  super-resolution}.
\newblock \bibinfo{journal}{\emph{arXiv preprint arXiv:1607.08569}}
  (\bibinfo{year}{2016}).
\newblock


\bibitem[\protect\citeauthoryear{Rist, Emmerichs, Enzweiler, and Gavrila}{Rist
  et~al\mbox{.}}{2021}]%
        {Rist2020SemanticSC}
\bibfield{author}{\bibinfo{person}{Christoph Rist}, \bibinfo{person}{David
  Emmerichs}, \bibinfo{person}{Markus Enzweiler}, {and} \bibinfo{person}{Dariu
  Gavrila}.} \bibinfo{year}{2021}\natexlab{}.
\newblock \showarticletitle{Semantic scene completion using local deep implicit
  functions on LiDAR data}.
\newblock \bibinfo{journal}{\emph{IEEE Transactions on Pattern Analysis and
  Machine Intelligence}} (\bibinfo{year}{2021}).
\newblock


\bibitem[\protect\citeauthoryear{Rist, Schmidt, Enzweiler, and Gavrila}{Rist
  et~al\mbox{.}}{2020}]%
        {Rist2020SCSSnetLS}
\bibfield{author}{\bibinfo{person}{Christoph~B Rist}, \bibinfo{person}{David
  Schmidt}, \bibinfo{person}{Markus Enzweiler}, {and} \bibinfo{person}{Dariu~M
  Gavrila}.} \bibinfo{year}{2020}\natexlab{}.
\newblock \showarticletitle{SCSSnet: Learning Spatially-Conditioned Scene
  Segmentation on LiDAR Point Clouds}. In \bibinfo{booktitle}{\emph{2020 IEEE
  Intelligent Vehicles Symposium (IV)}}. IEEE, \bibinfo{pages}{1086--1093}.
\newblock


\bibitem[\protect\citeauthoryear{Scharstein and Pal}{Scharstein and
  Pal}{2007}]%
        {Scharstein2007LearningCR}
\bibfield{author}{\bibinfo{person}{Daniel Scharstein} {and}
  \bibinfo{person}{Chris Pal}.} \bibinfo{year}{2007}\natexlab{}.
\newblock \showarticletitle{Learning conditional random fields for stereo}. In
  \bibinfo{booktitle}{\emph{2007 IEEE Conference on Computer Vision and Pattern
  Recognition}}. IEEE, \bibinfo{pages}{1--8}.
\newblock


\bibitem[\protect\citeauthoryear{Silberman, Hoiem, Kohli, and Fergus}{Silberman
  et~al\mbox{.}}{2012}]%
        {Silberman2012IndoorSA}
\bibfield{author}{\bibinfo{person}{Nathan Silberman}, \bibinfo{person}{Derek
  Hoiem}, \bibinfo{person}{Pushmeet Kohli}, {and} \bibinfo{person}{Rob
  Fergus}.} \bibinfo{year}{2012}\natexlab{}.
\newblock \showarticletitle{Indoor segmentation and support inference from rgbd
  images}. In \bibinfo{booktitle}{\emph{European conference on computer
  vision}}. Springer, \bibinfo{pages}{746--760}.
\newblock


\bibitem[\protect\citeauthoryear{Sitzmann, Martel, Bergman, Lindell, and
  Wetzstein}{Sitzmann et~al\mbox{.}}{2020}]%
        {Sitzmann2020ImplicitNR}
\bibfield{author}{\bibinfo{person}{Vincent Sitzmann}, \bibinfo{person}{Julien
  Martel}, \bibinfo{person}{Alexander Bergman}, \bibinfo{person}{David
  Lindell}, {and} \bibinfo{person}{Gordon Wetzstein}.}
  \bibinfo{year}{2020}\natexlab{}.
\newblock \showarticletitle{Implicit neural representations with periodic
  activation functions}.
\newblock \bibinfo{journal}{\emph{Advances in Neural Information Processing
  Systems}}  \bibinfo{volume}{33} (\bibinfo{year}{2020}).
\newblock


\bibitem[\protect\citeauthoryear{Sitzmann, Zollh{\"o}fer, and
  Wetzstein}{Sitzmann et~al\mbox{.}}{2019}]%
        {Sitzmann2019SceneRN}
\bibfield{author}{\bibinfo{person}{Vincent Sitzmann}, \bibinfo{person}{Michael
  Zollh{\"o}fer}, {and} \bibinfo{person}{Gordon Wetzstein}.}
  \bibinfo{year}{2019}\natexlab{}.
\newblock \showarticletitle{Scene representation networks: Continuous
  3d-structure-aware neural scene representations}.
\newblock \bibinfo{journal}{\emph{arXiv preprint arXiv:1906.01618}}
  (\bibinfo{year}{2019}).
\newblock


\bibitem[\protect\citeauthoryear{Song, Yu, Zeng, Chang, Savva, and
  Funkhouser}{Song et~al\mbox{.}}{2017}]%
        {song2017semantic}
\bibfield{author}{\bibinfo{person}{Shuran Song}, \bibinfo{person}{Fisher Yu},
  \bibinfo{person}{Andy Zeng}, \bibinfo{person}{Angel~X Chang},
  \bibinfo{person}{Manolis Savva}, {and} \bibinfo{person}{Thomas Funkhouser}.}
  \bibinfo{year}{2017}\natexlab{}.
\newblock \showarticletitle{Semantic scene completion from a single depth
  image}. In \bibinfo{booktitle}{\emph{Proceedings of the IEEE Conference on
  Computer Vision and Pattern Recognition}}. \bibinfo{pages}{1746--1754}.
\newblock


\bibitem[\protect\citeauthoryear{Su, Jampani, Sun, Gallo, Learned-Miller, and
  Kautz}{Su et~al\mbox{.}}{2019}]%
        {su2019pixel}
\bibfield{author}{\bibinfo{person}{Hang Su}, \bibinfo{person}{Varun Jampani},
  \bibinfo{person}{Deqing Sun}, \bibinfo{person}{Orazio Gallo},
  \bibinfo{person}{Erik Learned-Miller}, {and} \bibinfo{person}{Jan Kautz}.}
  \bibinfo{year}{2019}\natexlab{}.
\newblock \showarticletitle{Pixel-adaptive convolutional neural networks}. In
  \bibinfo{booktitle}{\emph{Proceedings of the IEEE/CVF Conference on Computer
  Vision and Pattern Recognition}}. \bibinfo{pages}{11166--11175}.
\newblock


\bibitem[\protect\citeauthoryear{Tang, Gao, and Hu}{Tang et~al\mbox{.}}{2021}]%
        {tang2021rgln}
\bibfield{author}{\bibinfo{person}{Jiaxiang Tang}, \bibinfo{person}{Xiang Gao},
  {and} \bibinfo{person}{Wei Hu}.} \bibinfo{year}{2021}\natexlab{}.
\newblock \showarticletitle{RGLN: Robust Residual Graph Learning Networks via
  Similarity-Preserving Mapping on Graphs}. In \bibinfo{booktitle}{\emph{ICASSP
  2021-2021 IEEE International Conference on Acoustics, Speech and Signal
  Processing (ICASSP)}}. IEEE, \bibinfo{pages}{2940--2944}.
\newblock


\bibitem[\protect\citeauthoryear{Tomasi and Manduchi}{Tomasi and
  Manduchi}{1998}]%
        {Tomasi1998BilateralFF}
\bibfield{author}{\bibinfo{person}{Carlo Tomasi} {and} \bibinfo{person}{Roberto
  Manduchi}.} \bibinfo{year}{1998}\natexlab{}.
\newblock \showarticletitle{Bilateral filtering for gray and color images}. In
  \bibinfo{booktitle}{\emph{Sixth international conference on computer vision
  (IEEE Cat. No. 98CH36271)}}. IEEE, \bibinfo{pages}{839--846}.
\newblock


\bibitem[\protect\citeauthoryear{Veli{\v{c}}kovi{\'c}, Cucurull, Casanova,
  Romero, Lio, and Bengio}{Veli{\v{c}}kovi{\'c} et~al\mbox{.}}{2017}]%
        {Velickovic2018GraphAN}
\bibfield{author}{\bibinfo{person}{Petar Veli{\v{c}}kovi{\'c}},
  \bibinfo{person}{Guillem Cucurull}, \bibinfo{person}{Arantxa Casanova},
  \bibinfo{person}{Adriana Romero}, \bibinfo{person}{Pietro Lio}, {and}
  \bibinfo{person}{Yoshua Bengio}.} \bibinfo{year}{2017}\natexlab{}.
\newblock \showarticletitle{Graph attention networks}.
\newblock \bibinfo{journal}{\emph{arXiv preprint arXiv:1710.10903}}
  (\bibinfo{year}{2017}).
\newblock


\bibitem[\protect\citeauthoryear{Wang, Sun, Liu, Sarma, Bronstein, and
  Solomon}{Wang et~al\mbox{.}}{2019}]%
        {Wang2019DynamicGC}
\bibfield{author}{\bibinfo{person}{Yue Wang}, \bibinfo{person}{Yongbin Sun},
  \bibinfo{person}{Ziwei Liu}, \bibinfo{person}{Sanjay~E Sarma},
  \bibinfo{person}{Michael~M Bronstein}, {and} \bibinfo{person}{Justin~M
  Solomon}.} \bibinfo{year}{2019}\natexlab{}.
\newblock \showarticletitle{Dynamic graph cnn for learning on point clouds}.
\newblock \bibinfo{journal}{\emph{Acm Transactions On Graphics (tog)}}
  \bibinfo{volume}{38}, \bibinfo{number}{5} (\bibinfo{year}{2019}),
  \bibinfo{pages}{1--12}.
\newblock


\bibitem[\protect\citeauthoryear{Weder, Schonberger, Pollefeys, and
  Oswald}{Weder et~al\mbox{.}}{2020}]%
        {weder2020routedfusion}
\bibfield{author}{\bibinfo{person}{Silvan Weder}, \bibinfo{person}{Johannes
  Schonberger}, \bibinfo{person}{Marc Pollefeys}, {and}
  \bibinfo{person}{Martin~R Oswald}.} \bibinfo{year}{2020}\natexlab{}.
\newblock \showarticletitle{Routedfusion: Learning real-time depth map fusion}.
  In \bibinfo{booktitle}{\emph{Proceedings of the IEEE/CVF Conference on
  Computer Vision and Pattern Recognition}}. \bibinfo{pages}{4887--4897}.
\newblock


\bibitem[\protect\citeauthoryear{Xing, Wang, Chen, and Zeng}{Xing
  et~al\mbox{.}}{2019a}]%
        {xing20192}
\bibfield{author}{\bibinfo{person}{Yajie Xing}, \bibinfo{person}{Jingbo Wang},
  \bibinfo{person}{Xiaokang Chen}, {and} \bibinfo{person}{Gang Zeng}.}
  \bibinfo{year}{2019}\natexlab{a}.
\newblock \showarticletitle{2.5 D convolution for RGB-D semantic segmentation}.
  In \bibinfo{booktitle}{\emph{2019 IEEE International Conference on Image
  Processing (ICIP)}}. IEEE, \bibinfo{pages}{1410--1414}.
\newblock


\bibitem[\protect\citeauthoryear{Xing, Wang, Chen, and Zeng}{Xing
  et~al\mbox{.}}{2019b}]%
        {xing2019coupling}
\bibfield{author}{\bibinfo{person}{Yajie Xing}, \bibinfo{person}{Jingbo Wang},
  \bibinfo{person}{Xiaokang Chen}, {and} \bibinfo{person}{Gang Zeng}.}
  \bibinfo{year}{2019}\natexlab{b}.
\newblock \showarticletitle{Coupling two-stream RGB-D semantic segmentation
  network by idempotent mappings}. In \bibinfo{booktitle}{\emph{2019 IEEE
  International Conference on Image Processing (ICIP)}}. IEEE,
  \bibinfo{pages}{1850--1854}.
\newblock


\bibitem[\protect\citeauthoryear{Xu, Wang, Ceylan, Mech, and Neumann}{Xu
  et~al\mbox{.}}{2019}]%
        {Xu2019DISNDI}
\bibfield{author}{\bibinfo{person}{Qiangeng Xu}, \bibinfo{person}{Weiyue Wang},
  \bibinfo{person}{Duygu Ceylan}, \bibinfo{person}{Radomir Mech}, {and}
  \bibinfo{person}{Ulrich Neumann}.} \bibinfo{year}{2019}\natexlab{}.
\newblock \showarticletitle{Disn: Deep implicit surface network for
  high-quality single-view 3d reconstruction}.
\newblock \bibinfo{journal}{\emph{arXiv preprint arXiv:1905.10711}}
  (\bibinfo{year}{2019}).
\newblock


\bibitem[\protect\citeauthoryear{Yang, Ye, Li, Hou, and Wang}{Yang
  et~al\mbox{.}}{2014}]%
        {Yang2014ColorGuidedDR}
\bibfield{author}{\bibinfo{person}{Jingyu Yang}, \bibinfo{person}{Xinchen Ye},
  \bibinfo{person}{Kun Li}, \bibinfo{person}{Chunping Hou}, {and}
  \bibinfo{person}{Yao Wang}.} \bibinfo{year}{2014}\natexlab{}.
\newblock \showarticletitle{Color-guided depth recovery from RGB-D data using
  an adaptive autoregressive model}.
\newblock \bibinfo{journal}{\emph{IEEE transactions on image processing}}
  \bibinfo{volume}{23}, \bibinfo{number}{8} (\bibinfo{year}{2014}),
  \bibinfo{pages}{3443--3458}.
\newblock


\bibitem[\protect\citeauthoryear{Yang, Yang, Davis, and Nist{\'e}r}{Yang
  et~al\mbox{.}}{2007}]%
        {Yang2007SpatialDepthSR}
\bibfield{author}{\bibinfo{person}{Qingxiong Yang}, \bibinfo{person}{Ruigang
  Yang}, \bibinfo{person}{James Davis}, {and} \bibinfo{person}{David
  Nist{\'e}r}.} \bibinfo{year}{2007}\natexlab{}.
\newblock \showarticletitle{Spatial-depth super resolution for range images}.
  In \bibinfo{booktitle}{\emph{2007 IEEE Conference on Computer Vision and
  Pattern Recognition}}. IEEE, \bibinfo{pages}{1--8}.
\newblock


\bibitem[\protect\citeauthoryear{Ye, Sun, Wang, Yang, Xu, Li, and Li}{Ye
  et~al\mbox{.}}{2020}]%
        {Ye2020DepthSV}
\bibfield{author}{\bibinfo{person}{Xinchen Ye}, \bibinfo{person}{Baoli Sun},
  \bibinfo{person}{Zhihui Wang}, \bibinfo{person}{Jingyu Yang},
  \bibinfo{person}{Rui Xu}, \bibinfo{person}{Haojie Li}, {and}
  \bibinfo{person}{Baopu Li}.} \bibinfo{year}{2020}\natexlab{}.
\newblock \showarticletitle{Depth Super-Resolution via Deep Controllable
  Slicing Network}. In \bibinfo{booktitle}{\emph{Proceedings of the 28th ACM
  International Conference on Multimedia}}. \bibinfo{pages}{1809--1818}.
\newblock


\bibitem[\protect\citeauthoryear{Zhang, Gool, and Timofte}{Zhang
  et~al\mbox{.}}{2020}]%
        {Zhang2020DeepUN}
\bibfield{author}{\bibinfo{person}{Kai Zhang}, \bibinfo{person}{Luc~Van Gool},
  {and} \bibinfo{person}{Radu Timofte}.} \bibinfo{year}{2020}\natexlab{}.
\newblock \showarticletitle{Deep unfolding network for image super-resolution}.
  In \bibinfo{booktitle}{\emph{Proceedings of the IEEE/CVF Conference on
  Computer Vision and Pattern Recognition}}. \bibinfo{pages}{3217--3226}.
\newblock


\bibitem[\protect\citeauthoryear{Zhang, Tian, Kong, Zhong, and Fu}{Zhang
  et~al\mbox{.}}{2018}]%
        {Zhang2018ResidualDN}
\bibfield{author}{\bibinfo{person}{Yulun Zhang}, \bibinfo{person}{Yapeng Tian},
  \bibinfo{person}{Yu Kong}, \bibinfo{person}{Bineng Zhong}, {and}
  \bibinfo{person}{Yun Fu}.} \bibinfo{year}{2018}\natexlab{}.
\newblock \showarticletitle{Residual dense network for image super-resolution}.
  In \bibinfo{booktitle}{\emph{Proceedings of the IEEE conference on computer
  vision and pattern recognition}}. \bibinfo{pages}{2472--2481}.
\newblock


\bibitem[\protect\citeauthoryear{Zhang}{Zhang}{2012}]%
        {zhang2012kinect}
\bibfield{author}{\bibinfo{person}{Zhengyou Zhang}.}
  \bibinfo{year}{2012}\natexlab{}.
\newblock \showarticletitle{Microsoft kinect sensor and its effect}.
\newblock \bibinfo{journal}{\emph{IEEE multimedia}} \bibinfo{volume}{19},
  \bibinfo{number}{2} (\bibinfo{year}{2012}), \bibinfo{pages}{4--10}.
\newblock


\end{thebibliography}
